\documentclass{tlp}
\usepackage{times} 
\usepackage[latin1]{inputenc}

\usepackage{latexsym}
\usepackage{graphicx}
\usepackage{url}
\usepackage{xspace}
\usepackage{aopmath}

  \renewenvironment{thebibliography}[1]{%
    \begin{oldthebibliography}{#1}%
      \setlength{\parskip}{0.1ex}%
      \setlength{\itemsep}{0.1ex}%
  }%
  {%
    \end{oldthebibliography}%
  }

\renewcommand{\paragraph}[1]{\medskip\noindent\textbf{#1\ \ }}

\newcommand{\nop}[1]{}
\def\punto{\hspace*{\fill}{\rule[0.5mm]{1.8mm}{1.8mm}}}

\newcommand{\NP}{{\rm NP}\xspace}
\newcommand{\SigmaP}[1]{{\Sigma}_{#1}^{P}}
\newcommand{\CONP}{\textrm{co-NP}\xspace}
\newcommand{\PiP}[1]{{\Pi}_{#1}^{P}}
\newcommand{\DeltaP}[1]{\ensuremath{\Delta_{#1}^P}}

\newcommand{\Or}{\ensuremath{\vee}}
\newcommand{\derives}{\mbox{\,:\hspace{-0.1em}\texttt{-}}\,\xspace}
\newcommand{\wderives}{\ensuremath{:\sim}}
\newcommand{\Comma}{\,,\ }

\newcommand{\WC}[1]{\ensuremath{WC(#1)}}
\newcommand{\Rules}[1]{\ensuremath{Rules(#1)}}

\newcommand{\p}{\ensuremath{{\cal P}}\xspace}
\newcommand{\n}{\ensuremath{{\cal N}}}

\newcommand{\Lg}{\ensuremath{\mathit{Lg}}}
\newcommand{\Rg}{\ensuremath{\mathit{Rg}}}
\newcommand{\GP}{\ensuremath{Ground(\p)}\xspace}
\newcommand{\BP}{\ensuremath{B_{\p}}\xspace}
\newcommand{\UP}{\ensuremath{U_{\p}}\xspace}
\newcommand{\UPn}{\ensuremath{U_{\p}^{\n}}\xspace}

\newcommand{\GR}{\ensuremath{Ground(\R)}}
\newcommand{\GRules}{\ensuremath{GroundRules(\p)}\xspace}
\newcommand{\GWC}{\ensuremath{GroundWC(\p)}}
\newcommand{\Gaweak}{\ensuremath{Ground(\aweak)}}

\newcommand{\wmax}[1]{\ensuremath{w_{max}^{#1}}}
\newcommand{\lmax}[1]{\ensuremath{l_{max}^{#1}}}

\newcommand{\R}{\ensuremath{r}}
\newcommand{\aweak}[1][]{\ensuremath{w_{#1}}}
\newcommand{\aprog}[1][]{\ensuremath{\p_{#1}}}
\newcommand{\HR}{\ensuremath{H(\R)}}
\newcommand{\BR}{\ensuremath{B(\R)}}
\newcommand{\BpR}{\ensuremath{B^{+}(\R)}}
\newcommand{\BnR}{\ensuremath{B^{-}(\R)}}

\newcommand{\defr}[1]{\ensuremath{def^r(#1)}}
\newcommand{\defp}[1]{\ensuremath{def^P(#1)}}

\newcommand{\dlfact}[1]{\ensuremath{#1.}}

\newcommand{\dlrule}[2]{\ensuremath{#1 \derives #2.}}

\newcommand{\dlweakconstraint}[3]{\ensuremath{\wderives #1.\ [#2:#3]}}

\newenvironment{simpleprogram}[1][]
   {\vspace{-0.5ex}\begin{itemize}\item[]
      \tt
      \begin{tabbing}
      \code{#1}\ \= \kill
   }
   {\end{tabbing}\end{itemize}\vspace{-3ex}}

\newcommand{\spi}[2][]{\code{#1}\>\code{#2}\\}

\newcommand{\multi}[1]{\ensuremath{\overline{2}^{#1}}}

\newcommand{\tneg}{\mbox{\texttt{-}}\xspace}
\newcommand{\naf}{\ensuremath{\mathit{not\ }}}
\newcommand{\countagg}{\ensuremath{\mathtt{\# count}}\xspace}
\newcommand{\sumagg}{\ensuremath{\mathtt{\# sum}}\xspace}
\newcommand{\minagg}{\ensuremath{\mathtt{\# min}}\xspace}
\newcommand{\maxagg}{\ensuremath{\mathtt{\# max}}\xspace}
\newcommand{\timesagg}{\ensuremath{\mathtt{\# times}}\xspace}
\newcommand{\anyagg}{\ensuremath{\mathtt{\# any}}\xspace}
\newcommand{\avgagg}{\ensuremath{\mathtt{\# avg}}\xspace}

\newcommand{\countaggBold}{\ensuremath{\mathtt{\bf\# count}}}
\newcommand{\sumaggBold}{\ensuremath{\mathtt{\bf\# sum}}}
\newcommand{\minaggBold}{\ensuremath{\mathtt{\bf\# min}}}
\newcommand{\maxaggBold}{\ensuremath{\mathtt{\bf\# max}}}
\newcommand{\timesaggBold}{\ensuremath{\mathtt{\bf\# times}}}

\def\P{\ensuremath{\mathcal{P}}}

\newcommand{\dlv}{{\sc DLV}\xspace}
\newcommand{\dlva}{{\sc DLV}\ensuremath{^\mathcal{A}}\xspace}
\newcommand{\DLP}{{\sc DLP}\xspace}
\newcommand{\DLPA}{{\sc DLP}\ensuremath{^\mathcal{A}}\xspace}

\newcommand{\smodelsa}{{\sc Smodels$^A$}}

\newcommand{\tuple}[1]{\langle#1\rangle}

\newtheorem{example}[theorem]{Example}

\newcommand{\code}[1]{\ensuremath{#1}}
\newenvironment{dlvcode}
  {
   \begin{displaymath}\hspace{-0.24cm}\begin{array}{l}}
  {\end{array}\end{displaymath}}

\newenvironment{dlvcodelong}[1]
  {
   \begin{displaymath}\begin{array}{@{\hspace*{\parindent}}#1@{\hspace*{10cm}}}}
  {\end{array}\end{displaymath}}

\submitted { 30 June 2007}
\revised { 28 December 2007 }
\accepted { 1 February 2008}

\title[Design and Implementation of Aggregate Functions in the DLV System]{Design and Implementation of Aggregate Functions in the DLV System\thanks{A preliminary version of this work appeared in the Proceedings of IJCAI-03.}}
\author[W. Faber et al.]
    {Wolfgang Faber, Gerald Pfeifer, Nicola Leone, Tina Dell'Armi, Giuseppe Ielpa\\
    Department of Mathematics, University of Calabria \\
    87036 Rende (CS), Italy\\
    \email{\{faber,gerald,leone,dellarmi,ielpa\}@mat.unical.it}
}

\date{}

\begin{document}

\maketitle

\begin{abstract}
Disjunctive Logic Programming (\DLP) is a very expressive formalism:
it allows for expressing every property of finite structures that is
decidable in the complexity class $\SigmaP{2} (=\!\!\NP^{\NP})$. Despite
this high expressiveness, there are some simple properties, often
arising in real-world applications, which cannot be encoded in a
simple and natural manner. Especially properties that require the use
of arithmetic operators (like sum, times, or count) on a set or multiset of elements, which
satisfy some conditions, cannot be naturally expressed in classic
\DLP.

To overcome this deficiency, we extend \DLP by aggregate functions in
a conservative way. In particular, we avoid the introduction of
constructs with disputed semantics, by requiring aggregates to be
stratified. We formally define the semantics of the extended language
(called \DLPA), and illustrate how it can be profitably used for
representing knowledge. Furthermore, we analyze the computational
complexity of \DLPA, showing that the addition of aggregates does not
bring a higher cost in that respect. Finally, we provide an
implementation of \DLPA in \dlv -- a state-of-the-art
\DLP system -- and report on experiments which confirm the usefulness
of the proposed extension also for the efficiency of computation.
\end{abstract}


\section{Introduction}

Disjunctive Logic Programs (\DLP) are logic programs where (non-monotonic)
negation may occur in the bodies, and disjunction may occur in the heads
of rules \cite{mink-82}.
This language is very expressive in a precise mathematical sense:
under the answer set semantics \cite{gelf-lifs-91}
it allows to express every property of finite structures that is
decidable in the complexity class $\SigmaP{2}(=\!\!\NP^{\NP})$ \cite{eite-etal-97f}.
Therefore, under widely believed assumptions,
\DLP is strictly more expressive than
{\em normal} ({\em disjunction-free}) logic programming,
whose expressiveness is limited to properties decidable in $\NP$,
and it can express problems which cannot be translated
to satisfiability of CNF formulas in polynomial time.
Importantly, besides enlarging the class of applications which can be
encoded in the language, disjunction often allows for representing
problems of lower complexity in a simpler and arguably more
natural fashion, cf.\ \cite{eite-etal-2000c}.

\paragraph{The problem.}
Despite this high expressiveness there are some simple properties,
often arising in real-world applications,
which cannot be encoded in \DLP in a simple and natural manner.
Among these are properties which require the application of arithmetic
operators such as count, sum, or min on a set of elements satisfying
some conditions.

Suppose, for instance, that you want to know if the sum of the
salaries of the employees working in a team exceeds a given budget
(see {Team Building} in Section~\ref{sec:KR}).  Using standard \DLP,
one first has to define an order over the employees, yielding a
successor relation. Then, one has to define a {\em sum} predicate in a
recursive way using this successor relation, computing the sum of all
salaries, and compare its result with the given budget. This approach
has two drawbacks: (1) It is bad from the KR perspective, as the
encoding is not immediate and not natural at all. In particular, an
ordering or successor relation often is not available and has to be
provided in an explicit manner.  (2) It is inefficient, as the
(instantiation of the) program is quadratic (in the cardinality of the
input set of employees).

Thus, there is a clear need to enrich \DLP
with suitable constructs for the natural representation of such properties
and to provide means for an efficient evaluation.

\paragraph{Contribution.}
We overcome the outlined deficiency of \DLP.  Instead of inventing new
constructs from scratch, as in some approaches in the literature (e.g., \cite{simo-etal-2002}), we
extend the language with aggregate functions, like those studied
in the context of databases, and implement them in
\dlv~\cite{leon-etal-2002-dlv} -- a state-of-the-art Disjunctive Logic
Programming system. The main advantages of this approach are that
extensibility of the language (both syntactically and semantically) is
straightforward, that aggregate functions are widely used, for
instance in database query languages, and that many issues arising from
the use of aggregates are well-understood.

The main contributions of this paper are the following:

\begin{itemize}
\item 
We extend Disjunctive Logic Programming by aggregate functions
and formally define the semantics of the resulting language, named \DLPA.
Actually, we introduce aggregates in the full \dlv language, that is, \DLPA includes also weak constraints \cite{bucc-etal-2000a}.
\item
We address knowledge representation issues, showing the impact of the
new constructs and describe ways how they can be employed
profitably on relevant problems.
We also highlight the usefulness of {\em assignment aggregates},
a new feature of  \DLPA, which is not supported by other
ASP systems with aggregates.
\item
We analyze the computational complexity of \DLPA.
We consider \DLPA programs with and without weak constraints.
Importantly, it turns out that in both cases
the addition of (stratified) aggregates does not increase
the computational complexity,
which remains the same as for reasoning on aggregate-free programs.
\item
We provide an implementation of \DLPA in the \dlv system,
deriving new algorithms and optimization techniques for
efficient evaluation.
\item
We report on experimentation, evaluating
the impact of the proposed language extension on efficiency.
The experiments confirm that, besides providing relevant advantages
from the knowledge representation point of view, aggregate functions
can bring significant computational gains.
\item
We compare \DLPA with related work proposed in the literature.
\end{itemize}

The result of this work is a concrete and powerful tool
for knowledge representation and reasoning, enhancing the modeling
features of standard DLP and Answer Set Programming (ASP) systems.

\DLPA, as described in this article, requires aggregates
to be stratified, that is, predicates defined by means of
aggregates are not allowed to mutually depend on each other. The
reason is that the set of stratified aggregate programs is the largest class
on which all major semantics proposed in the literature coincide.
Moreover, the introduction of unstratified aggregates causes a computational
overhead in some cases, while the computational complexity of the reasoning
tasks remains the same if stratified aggregates are introduced.
(See Section~\ref{sec:related-stratification} for a discussion about
this issue.)

It is worthwhile noting that, compared with other implementations of
aggregates in DLP and ASP, the language of our system supports some
extra features which turn out to be very useful in practice for KR
applications.  For instance, the {\em Fastfood problem}, described in
Section \ref{sec:KR}, is represented naturally and compactly in our
language, while its encoding in the language of other DLP and ASP
systems seems to be more involved causing computation to be
dramatically less efficient, due to their more severe safety
restrictions (domain predicates), and also to the lack of the ``min''
aggregate function (see Section \ref{sec:lparse}).

The paper is organized as follows.
Section \ref{sec:language} illustrates the \DLPA language,
providing a formal specification of both the syntax and the semantics
of our extension of \DLP with aggregates.
Section \ref{sec:KR} addresses knowledge representation issues,
showing the profitable employment of aggregate functions
in a couple of examples.
Section \ref{sec:complexity} analyzes the computational complexity of the
\DLPA language.
Section \ref{sec:implementation} addresses some implementation issues.
Section \ref{sec:experiments} reports on the results of the experimentation activity.
Section \ref{sec:related} discusses related works.
Finally, in Section \ref{sec:conclusion} we draw our conclusions.

\section{The \DLPA Language}
\label{sec:language}

In this section we provide a formal definition of the syntax and
semantics of the \DLPA language.
\DLPA is an extension of the language of the \dlv system
by set-oriented (or aggregate) functions.
Specifically, \DLPA includes disjunction, default (or non-monotonic)
negation, integrity and weak constraints, and aggregates.%
\footnote{We do not treat strong negation explicitly.
\dlv supports this by a simple rewriting technique, adding
a constraint $\derives a, \tneg a$ for each strongly negated atom
$\tneg a$, where $a$ also occurs in the program.}  For further
background we refer to \cite{gelf-lifs-91},
\cite{bara-2002}, and \cite{leon-etal-2002-dlv}.

\subsection{Syntax}

We assume sets of {variables}, {constants}, and
{predicates} to be given.  Similar to Prolog, we assume \emph{variables} to
be strings starting with uppercase letters and \emph{constants} to be non-negative
integers or strings starting with lowercase letters.  \emph{Predicates} are
strings starting with lowercase letters or symbols such as $=$, $<$, $>$ 
(so called built-in predicates that have a fixed meaning).
An \emph{arity} (non-negative integer) is associated with each predicate. 

\paragraph{Standard Atoms and Literals.}
A {\em term} is either a variable or a constant.
A {\em standard atom} is an expression $p(t_{1},\ldots$,$t_{n})$, where $p$ is
a {\em predicate} of arity $n$ and $t_{1}$,\ldots,$t_{n}$ are terms.
A {\em standard literal} $L$
is either a standard atom $A$ (in this case, it is {\em positive}) or a
standard atom $A$ preceded by the default negation symbol \naf
(in this case, it is {\em negative}).
A conjunction of standard literals is of the form $\mathit L_1,\ldots,L_k$ where
each $L_i$ ($1<=i<=k$) is a standard literal.

A structure (e.g.~standard atom, standard literal, conjunction) is
ground, if neither the structure itself nor any substructures contain
any variables.

\paragraph{Sets.}
A (\DLPA) {\em set} is either a symbolic set or a ground set.
A {\em symbolic set} is a pair \mbox{$\{\mathit{Vars}\! :\! \mathit{Conj}\}$},
where $\mathit{Vars}$ is a comma-separated list of variables
and $\mathit{Conj}$ is a conjunction of standard literals.
Intuitively, a symbolic set $\{X \!\! : \!\! a(X,Y),\naf p(Y)\}$
stands for the set of $X$-values making the conjunction
$a(X,Y),\naf p(Y)$ true, i.e.,
$\{X\! :\! \exists Y \mathit{such\ that}\ a(X,Y)\land \naf p(Y)\ \mathit{is\ true}\}$;
see Section~\ref{sec:semantics} for details.

A {\em ground set} is a set of pairs of the form
$\tuple{\overline{t}\! :\! \mathit{Conj}}$,
where $\overline{t}$ is a list of constants and $\mathit{Conj}$ is a ground
conjunction of standard literals.

\paragraph{Aggregate Functions and Aggregate Atoms.}
An {\em aggregate function} is of the form $f(S)$, where $S$ is a set, 
and $f$ is a {\em function name} among \countagg, \minagg, \maxagg, \sumagg,
\timesagg.  
An {\em aggregate atom} is
  \[Lg \prec_1 f(S) \prec_2 \Rg\]
where $f(S)$ is an aggregate function,
$\prec_1,\prec_2 \in \{ =,\ <,\ \leq, >, \geq \}$,
and $\Lg$ and $\Rg$ (called {\em left guard}, and {\em right guard},
respectively) are terms.
One of ``$\Lg \prec_1$'' and ``$\prec_2 \Rg$'' can be omitted.  In
this case, ``$0 \leq$'' and ``$\leq +\infty$'' are assumed,
respectively. If both $\prec_1,\prec_2$ are present, we assume for
simplicity that $\prec_1 \in \{<,\leq\}$ if and only if $\prec_2 \in
\{<,\leq\}$ and that both $\prec_1$ and $\prec_2$ are different from
$=$.\footnote{The aggregates not considered are of limited importance,
as they impose two upper or two lower guards, of which one will be
redundant.}

\begin{example}
The following are two aggregate atoms.
The latter contains a ground set and could be a ground instance of the former.
\begin{dlvcode}
\maxagg\{Z: r(Z), a(Z,V)\} > Y\\
\maxagg\{\tuple{2: r(2), a(2,x)},\tuple{2: r(2), a(2,y)}\} > 1
\end{dlvcode}
\end{example}

\paragraph{(General) Atoms, Literals and Rules.}
An {\em atom} is either a standard atom or an aggregate atom.

A {\em literal} $L$ is an atom $A$ (positive literal) or an atom $A$
preceded by the default negation symbol \naf (negative literal).
If $A$ is an aggregate atom, $L$ is an {\em aggregate literal}.

A {\em (\DLPA) rule} \R{} is a construct

\[
a_1\ \Or\ \cdots\ \Or\ a_n\ \derives\
        b_1,\ldots, b_k, b_{k+1}, \ldots, b_m.
\]

\noindent 
where $a_1,\ldots ,a_n$ are standard atoms, $b_1,\ldots,b_k$ are positive literals,
and $b_{k+1},\ldots,b_m$ are negative literals,
and $n\geq 0$, $m\geq k \geq 0$, $m+n \geq 1$.
The disjunction $a_1\ \Or\ \cdots\ \Or\ a_n$ is the \emph{head} of \R{},
while the conjunction $b_1,\ldots,b_m$ is the \emph{body} of \R{}, 
$b_1,\ldots,b_k$ being the \emph{positive body} and $b_{k+1},\ldots,b_m$ the \emph{negative body}.
We define $\HR = \{a_1,\ldots,a_n\}$, $\BR = \{b_1,\ldots,b_m\}$, $\BpR$
$ = \{b_1,\ldots,b_k\}$, and $\BnR = \{b_{k+1}, \ldots, b_m\}$.
A rule without head literals (i.e.\ $n=0$) is usually referred to as an {\em integrity constraint}.      
A rule with an empty body (i.e.\ $m=0$) is called a fact, and we usually
omit the ``\derives{}'' sign in this case.

\paragraph{Weak Constraints.}
The language of \dlv, that we enhance by aggregates in this paper,
extends disjunctive Datalog by another construct called
{\em weak constraint} \cite{bucc-etal-2000a}.
The \DLPA language allows for a general form of weak constraints
also including aggregate literals.

We define weak constraints as a variant of integrity constraints. In
order to differentiate between these two, weak constraints use the symbol
``\wderives{}'' instead of
``\derives{}''.
In addition, a weight and a priority level inducing a partial order
among weak constraints are specified.

Formally, a weak constraint $wc$ is an expression of the form
\[
\wderives\ b_1,\ldots, b_k, b_{k+1},\ldots, b_m.\ [w:l]
\]
where $b_1,\ldots,b_k$ are positive literals, $b_{k+1},\ldots,b_m$ are
negative literals, and $w$
(the \emph{weight}) and $l$ (the \emph{level}, or \emph{layer}) are
positive integer constants or variables.
For convenience, $w$, $l$, or both can be omitted and default to $1$
in this case.

\paragraph{\DLPA Programs.}
A {\em (\DLPA) program} \p{} ({\em program}, for short)
is a set of \DLPA rules (possibly including integrity constraints)
and weak constraints.
For a program \p, let \Rules{\p} denote the set of rules (including
integrity constraints), and let \WC{\p} denote the set of weak constraints
in \p{}.
A program is \emph{positive} if it does not contain any negative literal.

\subsection{Syntactic Restrictions and Notation}
\label{sec:restrictions}

We begin with two notions of stratification, which make use of the
concept of a level mapping. Functions $||\ ||$ from predicates in a
program $\P$ to finite ordinals are called {\em level mappings} of
$\P$.

\paragraph{Negation-stratification.}

A program  $\P$
is called {\em negation-stratified} \cite{apt-etal-88,przy-88},
if there is a level mapping
$||\ ||_n$ of $\P$ such that, for each pair $p$ and $p'$ of predicates of $\P$ and every rule $r$ of $\P$,
\begin{enumerate}
\item if $p$ occurs in $\BpR$ and $p'$ occurs in $\HR$, then
$||p||_n \leq ||p'||_n$; and
\item if $p$ occurs in $\BnR$ and $p'$ occurs in $\HR$, then
$||p||_n < ||p'||_n$; and
\item if $p$ and $p'$ occur in $\HR$, then $||p||_n = ||p'||_n$. 
\end{enumerate}

\paragraph{Aggregate-stratification.}

The idea of aggregate-stratification is that two predicates defined
by means of aggregates do not mutually depend on one another.

A \DLPA program \P\ is {\em aggregate-stratified} if there exists
a level mapping $||\ ||_a$
such that for each pair $p$ and $p'$ of predicates
of \P, and for each rule $\R \in \P$,
\begin{enumerate}
\item if $p$ occurs in a standard atom in $\BR$ and $p'$ occurs in $\HR$,
then $||p||_a\leq ||p'||_a$; and
\item if $p$ occurs in an aggregate atom in $\BR$, and $p'$ occurs in $\HR$,
then $||p||_a < ||p'||_a$; and
\item if $p$ and $p'$ occur in $\HR$, then $||p||_a = ||p'||_a$. 
\end{enumerate}

\begin{example}\label{ex:stratification}
Consider a program consisting of a set of facts for predicates $a$
and $b$, plus the following two rules:

\begin{dlvcode}
q(X) \derives p(X), \countagg\{Y : a(Y,X),b(X)\} \leq 2. \\
p(X) \derives q(X), b(X).\\
\end{dlvcode}
The program is aggregate-stratified, as the level mapping \ \
$||a||=||b||=1\quad ||p||=||q||=2$ \ \ satisfies the required
conditions.  If we add the rule $b(X) \derives p(X)$, no such level mapping
exists and the program becomes aggregate-unstratified, as in this case
a level mapping would have to satisfy $||q|| > ||b|| \geq ||p|| \geq
||q||$, hence $||q|| > ||q||$.
\punto\end{example}

Intuitively, aggregate-stratification forbids recursion through
aggregates.  It guarantees that the semantics of
aggregates is agreed upon and coherent with the intuition, while
the semantics of aggregate-unstratified programs is debatable, and
some semantic properties (like, e.g., existence of answer sets for
positive programs) are usually lost. For a more detailed discussion,
see Section~\ref{sec:related-stratification}.

\paragraph{Local and global variables, Safety.}
For simplicity, and without loss of generality,
we assume that the body of each rule and weak constraint
contains at most one aggregate atom.\footnote{Note that we do this
only to simplify the definitions; our implementation can deal with
multiple aggregates in one rule.}
A {\em local} variable of a rule $r$ is a variable appearing solely in
an aggregate function in $r$; a variable of $r$ which is not local
is called {\em global}.
A {\em nested} atom of $r$ is an atom appearing in an aggregate
atom of $r$; an atom of $r$ which is not nested is called {\em unnested}.

A rule or weak constraint $r$ is {\em safe} if the following
conditions hold: (i) each global variable of $r$ appears in a positive
unnested standard literal of the body of $r$; (ii) each local variable of $r$
that appears in a symbolic set $\{\mathit{Vars}:\mathit{Conj}\}$ also
appears in a positive literal in $\mathit{Conj}$.  Finally, a program
is safe if all of its rules and weak constraints are safe.%
\footnote{Note that the safety restrictions apply also to aggregate-free
rules and constraints.}

Condition (i) is the standard safety condition adopted in Datalog, to
guarantee that the variables are range restricted \cite{ullm-89},
while condition (ii) is specific to aggregates.

\begin{example}\label{ex:safety}
Consider the following rules:
\begin{dlvcode}
p(X) \derives q(X,Y,V), Y \!< \maxagg\{Z: r(Z),\naf a(Z,V)\}. \\
p(X) \derives q(X,Y,V), Y \!< \sumagg\{Z: \naf a(Z,S)\}. \\
p(X) \derives q(X,Y,V), T \!< \minagg\{Z: r(Z), \naf a(Z,V)\}.
\end{dlvcode}
The first rule is safe, while the second is not, since both local
variables $Z$ and $S$ violate condition (ii). The third rule is not
safe either, since the global variable $T$ violates condition (i).
\punto\end{example}

We assume in the following that \DLPA
programs are safe and aggregate-stratified, unless explicitly stated
otherwise.

\subsection{Semantics}
\label{sec:semantics}

Let us first define some notation which is to be used subsequently.
Given a \DLPA program \p, let \UP denote the set of constants appearing in \p,
$\UPn \subseteq \UP$ the set of the natural numbers occurring in \UP,
and \BP the set of standard atoms
constructible from the (standard) predicates of \p with constants in \UP.
Given a set $X$, let $\multi{X}$ denote the set of all
multisets over elements from $X$.

Let us next describe the domains and the meanings of the aggregate functions
considered in this work:

\begin{itemize}
\item[]
\countaggBold:  defined over $\multi{\UP}\!\!$, the number of elements in the set.\\
\sumaggBold:  defined over $\multi{\UPn}\!\!$, the sum of the numbers in the set; 0 in case of the empty set.\\
\timesaggBold:  over $\multi{\UPn}\!\!\!$, the product of the numbers
in the set; 1 for the empty set.\\
\minaggBold, \maxaggBold:  defined over $\multi{\UPn}\!-\{\emptyset\}$,
the minimum/maximum element in the set.%
\footnote{Note that \minagg\ and \maxagg\ can be easily extended
to the domain of the strings by considering the lexicographic ordering.}
\end{itemize}

If the argument of an aggregate function does not belong to its domain,
the aggregate evaluates to false (denoted as $\bot$).

\paragraph{Instantiation.}
A {\em substitution} is a mapping
from a set of variables to the set $\UP$ of the constants in \p.
A substitution from the set of global variables of a rule
$r$ (to $\UP$) is a {\em global substitution for r};
a substitution from the set of local variables of a symbolic set $S$ (to $\UP$)
is a {\em local substitution for $S$}.
Given a symbolic set without global variables
$S = \{\mathit{Vars}:\mathit{Conj}\}$,
the \emph{instantiation} of $S$ is the following ground set of pairs
$inst(S)$:
$\{ \tuple{\gamma(\mathit{Vars}): \gamma(\mathit{Conj})} \mid$
$\gamma$ {\em is a local substitution for }$S\}$.%
\footnote{Given a substitution $\sigma$ and a \DLPA object $O$
(rule, conjunction, set, etc.), we denote by $\sigma(O)$
the object obtained by replacing each variable $X$ in $O$ by $\sigma(X)$.}

A {\em ground instance} of a rule or a weak constraint $r$
is obtained in two steps:
(1) a global substitution $\sigma$ for $r$ is applied to $r$; and
(2) every symbolic set $S$ in $\sigma(r)$ is then replaced by its
instantiation $inst(S)$.
The instantiation \GP\ of a program \p is the set of all possible instances 
of the rules and the weak constraints of \p.
\begin{example}
\label{ex:instantiation}
Consider the following program $\p_1$:
\begin{dlvcode}
q(1) \Or p(2,2). \quad\quad\quad q(2) \Or p(2,1).\\
t(X) \derives q(X), \sumagg\{Y\!:p(X,Y)\}>1.
\end{dlvcode}
The instantiation $Ground(\p_1)$ is the following:
\begin{dlvcode}
q(1) \Or p(2,2). \quad\quad\quad q(2) \Or p(2,1). \\
t(1) \derives q(1), \sumagg\{\tuple{1:p(1,1)},\tuple{2:p(1,2)}\}>1. \\
t(2) \derives q(2), \sumagg\{\tuple{1:p(2,1)},\tuple{2:p(2,2)}\}>1.
\end{dlvcode}
\punto\end{example}

For any program \p{}, \GP{} denotes the set \( \displaystyle\GRules \cup \GWC,\) where
\( \displaystyle \GRules= \!\!\!\bigcup_{\R \in\Rules{\p}} \!\!\!\! \GR \)\ and\ \( \displaystyle \GWC = \!\!\!\bigcup_{\aweak \in\WC{\p}} \!\!\!\! \Gaweak \).
Note that for propositional programs, \( \p{} = \GP{} \) holds.

\paragraph{Interpretations and Models.}
An {\em interpretation} of a \DLPA program \p is a set of standard ground atoms
$I\subseteq \BP$.
The truth valuation $I(A)$, where $A$ is a standard ground literal or a standard ground conjunction,
is defined in the usual way.
Besides assigning truth-values to standard ground literals, an interpretation
provides meaning also to ground sets, aggregate functions and aggregate
literals;
the meaning of a set, an aggregate function, and an aggregate atom
under an interpretation, is a multiset, a value, and a truth-value,
respectively.
Let $f(S)$ be a a ground aggregate function. The valuation of the (ground) set $S$
w.r.t.\ $I$ is the multiset $I(S)$ defined as follows:
Let $S_I=\{\tuple{t_1,...,t_n}\mid
\tuple{t_1,...,t_n\!:\!\mathit{Conj}}\!\in\! S \,\wedge\,
\mathit{Conj}\textrm{ is true w.r.t.\ } I\}$, then $I(S)$
is the multiset obtained as the projection of the tuples of $S_I$ on
their first constant, that is
$I(S)=[t_1 \mid \tuple{t_1,...,t_n} \in S_I ]$.

The valuation $I(f(S))$ of a ground aggregate function $f(S)$ w.r.t.\ $I$
is the result of the application of $f$ on $I(S)$.
If the multiset $I(S)$ is not in the domain of $f$,
$I(f(S))= \bot$ (where $\bot$ is a fixed symbol not occurring in \p).

A ground aggregate atom $A = \Lg \prec_1 f(S) \prec_2 \Rg$ is {\em
true w.r.t.\ $I$} if: (i) $I(f(S))\neq \bot$, and (ii) the
relationships $\Lg \prec_1 I(f(S))$ and $I(f(S)) \prec_2 \Rg$
hold\footnote{Note that in the implemented system
(cf.~Section~\ref{sec:implementation}) an error will be produced if
$\Lg$ or $\Rg$ are not in $\UPn$.}; otherwise, $A$ is false.

\begin{example}
Let $I$ be the interpretation
$\{f(1), g(1,2), g(1,3), g(1,4), g(2,4), h(2), h(3), h(4)\}$.
With respect to the interpretation $I$, and assuming that all variables are local, we have that:

\medskip
\begin{itemize}
\item[-]
$\countagg\{ X : g(X,Y) \} > 2$ is false, because $S_I$ for the corresponding ground set is $\{\tuple{1},\tuple{2}\}$, so $I(S) = [1,2]$ and $\countagg([1,2])=2$.
\item[-]
$\countagg\{ X,Y : g(X,Y) \} > 2$ is true, because $S_I = \{\tuple{1,2},\tuple{1,3},\tuple{1,4},\tuple{2,4}\}$, $I(S) = [1,1,1,2]$ and $\countagg([1,1,1,2])=4$. 
\item[-]
$23 < \timesagg\{ Y : f(X), g(X,Y) \} <= 24$ is true; in this case $S_I = \{\tuple{2},\tuple{3},\tuple{4}\}$, $I(S) = [2,3,4]$ and $\timesagg([2,3,4])=24$.
\item[-]
$\sumagg\{ A : g(A,B), h(B) \} <= 3$ is true, as we have that $S_I = \{\tuple{1},\tuple{2}\}$, $I(S) = [1,2]$ and $\sumagg([1,2])=3$.
\item[-]
$\sumagg\{ A,B : g(A,B), h(B) \} <= 3$ is false, since $S_I = \{\tuple{1,2},\tuple{1,3},\tuple{1,4},\tuple{2,4}\}$, $I(S) = [1,1,1,2]$ and $\sumagg([1,1,1,2])=5$.;
\item[-]
$\minagg\{ X : f(X), g(X) \} >= 2$ is false
because the evaluation of (the instantiation of) $\{ X : f(X), g(X) \}$
w.r.t.\ $I$ yields the empty set,
which does not belong to the domain of $\minagg$
\ (we have that $I(\minagg\{\})=\bot$).
\end{itemize}
\punto\end{example}

A ground rule $r\in \GRules$ is {\em satisfied w.r.t.\ $I$} if some head atom
is true w.r.t.\ $I$ whenever all body literals are true w.r.t.\ $I$.
(If $r$ is an integrity constraint, $r$ is satisfied iff
its body is false.)
A ground weak constraint $w \in \GWC$ is {\em satisfied w.r.t.\ $I$} if some
body literal of $w$ is false w.r.t.\ $I$.

A {\em model} of \p is an interpretation $M$ of \p such that
every rule $\R \in$  $\GRules$ is satisfied w.r.t.\ $M$.  A model $M$ of 
\p is (subset) \emph{minimal} if no model $N$ of \p exists such that $N$
is a proper subset of $M$.%
\footnote{Note that a model can violate weak constraints.}

\begin{example}
\label{ex:semantics}
Consider the aggregate atom 
$A = \sumagg\{\tuple{1\!:\!p(2,1)},\tuple{2\!:\!p(2,2)}\}>1$
from Example~\ref{ex:instantiation}.
Let $S$ be the ground set appearing in $A$.
For the interpretation $I=\{q(2),p(2,2),t(2)\}$, $I(S)=[2]$, 
the application of \sumagg\ over $[2]$ yields $2$,
and $A$ is therefore true w.r.t.\ $I$, since $2>1$.
Indeed, one can verify that $I$ is a minimal model of the program of
Example~\ref{ex:instantiation}.
\punto\end{example}

\paragraph{Answer Sets.}
We define the answer sets of \DLPA programs in three steps, using
their ground instantiation.
First we define the answer sets of standard positive programs
(i.e., programs without aggregates and without weak constraints),
then we give a reduction of \DLPA programs
containing aggregates and negation as failure
to standard positive ones and use it to define answer sets of arbitrary
sets of rules, possibly containing aggregates and negation as failure.
Finally, we specify how weak constraints affect the semantics,
arriving at the semantics of general \DLPA programs
(with negation, aggregates and weak constraints).

\paragraph{Step 1}
An interpretation $X \subseteq \BP$ is an {\em answer set} of a
standard positive \DLPA program (without aggregates and weak
constraints) \p, if it is a minimal model of \p.

\begin{example}
The positive program $\p_1 =\{ a \Or{} b \Or{} c.\}$
has the answer sets $\{{a}\}$, $\{{b}\}$, and
$\{{c}\}$. Its extension $\p_2 =\{ a \Or{} b \Or{} c.\Comma \derives{} a.\}$
has the answer sets $\{b\}$ and $\{c\}$.
Finally, the positive program
$\p_3 = \{ a \Or{} b \Or{} c.\Comma \derives{} a.\Comma
  b \derives{} c.\Comma
  c \derives{} b.\}
$
has the single answer set $\{{b},{c}\}$.
\punto\end{example}

\paragraph{Step 2}
The {\em reduct} or {\em Gelfond-Lifschitz transform} of a
\DLPA program \p{} w.r.t.\ a set $X \subseteq \BP$ is the standard
positive ground program $\p^X$ obtained from \GRules by
\begin{itemize}
\item deleting all rules $\R \in \GRules$ for which a negative literal in
$\BR$ is false w.r.t.\ $X$ or an aggregate literal is false w.r.t.\ $X$; and
\item deleting all negative literals and aggregate literals
from the remaining rules.
\end{itemize}
An answer set of a program \p{} is a set $X \subseteq \BP$ such
that $X$ is an answer set of $\p^X$.

\begin{example}
Given the following aggregate-stratified program with negation
$\p_4 =$
\begin{dlvcode}
 \{ d(1). \Comma a \Or{} b \derives{} c.\Comma\\
  b \derives{} \naf a, \naf c,\countagg\{Y:d(Y)\}>0.\Comma\\
  a \Or{} c \derives{} \naf b, \sumagg\{Y:d(Y)\}>1. \}
\end{dlvcode}
and $I = \{ b, d(1)\}$, the reduct $\p_4^I$ is
$\{ d(1).\Comma a \Or{} b \derives{} c.\Comma b. \}$.
It is easy to see that $I$ is an answer set of $\p_4^I$, and
thus an answer set of $\p_4$ as well.

Now consider $J = \{a, d(1)\}$. The reduct $\p_4^J$ is
$\{d(1).\Comma a \Or{} b \derives{} c.\}$.
It can be easily verified that $J$ is a model of $\p_4^J$.  However,
also $J' = \{d(1)\} \subset J$ is a model of $\p_4^J$, so $J$ is not
an answer set of $\p_4^J$ and thus $J$ is not an answer set of $\p_4$.

For $K = \{c, d(1)\}$, on the other hand, the reduct
$\p_4^K$ is equal to
$\p_4^J$, but $K$ is not an answer set of $\p_4^K$: for the rule
$\R{}: a \Or{} b \derives{} c$,
$\BR \subseteq K$ holds, but $\HR \cap K
\neq \emptyset$ does not. Indeed, it can be verified that $I$ and $J$
are the only answer sets of $\p_4$.
\punto\end{example}

\paragraph{Step 3}
Given a ground program \p{} with weak constraints $WC(\p)$, we are
interested in the answer sets of $\Rules{\p}$ which minimize the sum
of weights of the violated (unsatisfied) weak constraints in the
highest priority level,%
\footnote{Higher values for weights and priority levels
mark weak constraints of higher importance.
The most important constraints are those having the highest
weight among those with the highest priority level.}
and among them those which minimize the sum of
weights of the violated weak constraints in the next lower level,
etc. Formally, this is expressed by an objective function $H^{\aprog{}}(A)$ for
\aprog{} and an
answer set $A$ as follows, using an auxiliary function $f_{\aprog{}}$
which maps leveled weights to weights without levels:
\[
\begin{array}{l@{\ }c@{\ }l}
f_{\aprog{}}(1) & = & 1,\\
f_{\aprog{}}(n) & = & f_{\aprog{}}(n-1) \cdot |\WC{\aprog{}}| \cdot \wmax{\aprog{}} + 1,\quad n > 1,\\
H^{\aprog{}}(A) & = & \sum_{i=1}^{\lmax{\aprog{}}} ( f_{\aprog{}}(i) \cdot \sum_{w \in N^{\aprog{}}_i(A)} weight(w) ),
\end{array}
\]
where \wmax{\aprog{}} and \lmax{\aprog{}} denote the maximum weight and maximum
level over the weak constraints in \aprog{}, respectively, $N^{\aprog{}}_i(A)$
denotes the set of the weak constraints in level $i$ that are violated by $A$,
and $weight(w)$ denotes the weight of the weak constraint $w$. Note that $|\WC{\aprog{}}|
\cdot \wmax{\aprog{}} + 1$ is greater than the sum of all weights in the
program, and therefore guaranteed to be greater than the sum of
weights of any single level.

Intuitively, the function $f_{\aprog{}}$ handles priority levels.
It guarantees that the violation of a single constraint of priority level $i$
is more ``expensive'' then the violation of \emph{all} weak constraints of the
lower levels (i.e., all levels $< i$).

 For a \DLPA program  \aprog{} (possibly with weak constraints),
a set $A$ is an {\em (optimal) answer set} of \aprog{} if and only if
(1) $A$ is an answer set of $\Rules{\aprog{}}$ and
(2) $H^{\aprog{}}(A)$ is minimal over all the answer sets of \Rules{\aprog{}}.

\begin{example} \label{exa:artificialweak}

Consider the following program $\p_5$, which has three weak
constraints:

\begin{simpleprogram}
\spi{\dlfact{a \Or b}}
\spi{\dlfact{b \Or c}}
\spi{\dlrule{d \Or nd\;}{\; a, c}}
\spi{\dlweakconstraint{\sumagg\{\tuple{4:b}\}>3 }{1}{2}}
\spi{\dlweakconstraint{a, nd}{4}{1}}
\spi{\dlweakconstraint{c, d}{3}{1}}
\end{simpleprogram}

$\Rules{\p_5}$ admits three answer sets: $A_1=\set{\code{a},
\code{c}, \code{d}}$, $A_2=\set{\code{a}, \code{c}, \code{nd}}$,
and $A_3=\set{\code{b}}$.  We have: $H^{\p_5}(A_1)=3$,
$H^{\p_5}(A_2)=4$, $H^{\p_5}(A_3)=13$. Thus, the unique 
(optimal) answer set is $\set{\code{a}, \code{c}, \code{d}}$ with
weight~3 in level~1 and weight~0 in level~2.
\end{example}

\subsection{Computing New Values from Aggregates}
\label{sec:assignmentpredicate}

Due to the definition of safety in Section~\ref{sec:restrictions} we
could define the semantics of aggregates using the standard Herbrand
Base and Herbrand Universe.
The values returned by aggregate functions do not extend the
Herbrand Universe.

This restriction, which is also imposed in the language of
the Lparse system \cite{lparse-manual} (see also Section~\ref{sec:lparse}), appears to be severe and limits the
expressiveness of the language.  Suppose, for instance, that the
employees of a company are stored by a number of facts of the form
$\mathit{employee}(Id,Name,Salary)$.  If the boss wants to know the
sum of the salaries she pays, a rule
\[
total(T) \derives T = \sumagg\{ S,I: \mathit{employee}(I,N,S)\}.
\]
would be most intuitive.%
\footnote{We aggregate over I (in addition to S), as otherwise
two employees having the same salary would count only once
in the total.  This is also why we allow for multisets.}

However, the above rule is unsafe because of the variable $T$.  Our
language thus fails to naturally express a simple query which can be
easily stated in SQL\footnote{Note that also the language of Lparse
cannot express this query, cf.\ Section~\ref{sec:lparse}.}.  To
overcome this problem, we introduce the notion of \emph{assignment
aggregate} and make appropriate adjustments to the notion of safety
and semantics.

\paragraph{Assignment Aggregate.}
We denote by $\defr{p}$ the set of \emph{defining rules} of a
predicate $p$, that is, those rules $r$ in which $p$ occurs in the
head. Moreover, the \emph{defining program} of a predicate $p$,
denoted  by $\defp{p}$, consists of $\defr{p}$ and the defining programs
of all predicates which occur in the bodies of rules in $\defr{p}$.

An aggregate atom is an \emph{assignment aggregate} if it is of the
form $X = f(S)$, $f(S) = X$, or $X = f(S) = X$, where $X$ is a
variable and for each predicate $p$ in $S$, $\defp{p}$ is
negation-stratified and non-disjunctive.

The intuition of the restriction on the definition of the nested
predicates is to ensure that these predicates are deterministically
computable.

\paragraph{Relaxed Safety.}
We slightly relax the notion of safety as defined in
Section~\ref{sec:restrictions}, changing only condition (i):

A rule or weak constraint $r$ is {\em safe} if the following
conditions hold: (i) each global variable of $r$ appears in a positive
unnested standard literal of the body of $r$ \emph{or as a guard of an assignment
aggregate}; (ii) each local variable of $r$ that appears in a symbolic
set $\{\mathit{Vars}:\mathit{Conj}\}$ also appears in a positive
literal in $\mathit{Conj}$.  Finally, a program is safe if all of its
rules and weak constraints are safe.

To adapt the formal semantics to this extension, we enrich the
Universe $\UP$ of the program by the set of positive integers which
result from the evaluation of an aggregate function, with a consequent
enlargement of $\BP$. Note that the (relaxed) safety criterion
guarantees domain independence of rules and weak constraints,
which---together with aggregate stratification---guarantees a simple
(and finite) evaluation. None of the remaining semantic notions needs
further adaptations.

\section{Knowledge Representation in \DLPA}
\label{sec:KR}
In this section, we show how aggregate functions can be used to encode
several relevant problems: Team Building, Seating, and a logistics
problem, called Fastfood.
Moreover, we show how some properties of the input relations
(e.g., the cardinality) can be simply computed
by using aggregates, and we describe the encoding of a variant
of the Fastfood problem.

\paragraph{Team Building.}
A project team has to be built from a set of employees
according to the following specifications:

\newcommand{\cmpenum}[2]{\item[]{\ensuremath{(#1_{#2})}}}
\begin{itemize}
\cmpenum{p}{1} 
The team consists of a certain number of employees.
\cmpenum{p}{2}
At least a given number of different skills must be present in the team.
\cmpenum{p}{3}
The sum of the salaries of the employees working in the team must not
exceed the given budget.
\cmpenum{p}{4}
The salary of each individual employee is within a specified
limit.
\cmpenum{p}{5}
The number of women working in the team has to reach at least a given number.
\end{itemize}

Information on our employees is provided by a number of facts
of the form\linebreak[4]
$emp(EmpId,Sex,Skill,Salary)$.
The size of the team, the minimum number of different skills in the
team, the budget, the maximum salary,
and the minimum number of women are specified by the facts 
$nEmp(N)$, $nSkill(N)$, $budget(B)$, $maxSal(M)$, and $women(W)$.
We then encode each property $p_i$ above by an aggregate atom $A_i$,
and enforce it by an integrity constraint containing $\naf A_i$.

\begin{dlvcodelong}{l}
\mbox{\% Guess whether to take an employee or not.}\\
in(I) \Or out(I) \derives emp(I,Sx,Sk,Sa).\\
\mbox{\% The team consists of exactly $N$ employees. ($p_1$)}\\
\derives nEmp(N), \naf \countagg\{I : in(I)\}=N.\\
\mbox{\% Overall, employees need to have at least $M$ different skills. ($p_2$)}\\
\derives nSkill(M), \naf \countagg\{Sk:emp(I,Sx,Sk,Sa),in(I)\}\geq M.\\
\mbox{\% The sum of the individual salaries must not exceed the budget $B$. ($p_3$)}\\
\derives budget(B), \naf \sumagg\{Sa,I:emp(I,Sx,Sk,Sa),in(I)\}\leq B.\\
\mbox{\% The max.\ salary in the team must not exceed the max.\ allowed salary $M$. ($p_4$)}\\
\derives maxSal(M), \naf \maxagg\{Sa:emp(I,Sx,Sk,Sa),in(I)\}\leq M.\\
\mbox{\% We have at least $W$ women in the team. ($p_5$)}\\
\derives women(W), \naf \countagg\{I:emp(I,f,Sk,Sa),in(I)\}\geq W.
\end{dlvcodelong}

Intuitively, the disjunctive rule ``guesses'' whether an employee is 
included in the team or not, while
the five constraints correspond one-to-one to the five requirements
$p_1$-$p_5$.
Thanks to the aggregates
the translation of the specification is surprisingly straightforward.

The example highlights the usefulness of representing both sets and multisets
in our language; the latter can be obtained by specifying more than
one variable in the $\mathit{Vars}$ part of a symbolic set
$\{\mathit{Vars}:\mathit{Conj}\}$).
For instance, the encoding of $p_2$ requires a \underline{set},
as we want to count {\em different} skills:
two employees in the team having the same skill count
once w.r.t.\ $p_2$.
On the contrary, $p_3$ requires to sum the elements of a 
\underline{multiset}:
if two employees have the same salary,
\emph{both} salaries should be summed up for $p_3$.
This is obtained by adding the variable $I$, which uniquely identifies
every employee, to $\mathit{Vars}$.
The valuation of \mbox{$\{Sa,I:emp(I,Sx,Sk,Sa),in(I)\}$} gives rise to the set
$S=\{\tuple{Sa,I}:Sa\ \mathit{is\ the\ salary\ of\ employee}$ $I$ $\mathit{in\ the\ team}\}$.
The sum function is then applied on the multiset of the first component
$Sa$ of all the tuples $\tuple{Sa,I}$ in $S$ (see Section~\ref{sec:semantics}).

\paragraph{Seating.}
We have to generate a seating arrangement for $k$ guests,
with $m$ tables and $n$ chairs per table.
Guests who like each other should sit at the same table;
guests who dislike each other should sit at different tables.

Suppose that the number of chairs per table is specified by
$nChairs(X)$ and that $person(P)$ and $table(T)$
represent the guests and the available tables, respectively.
Then, we can generate a seating arrangement by the following program:
\begin{dlvcodelong}{l}
\mbox{\% Guess whether person P sits at table T or not.}\\
at(P,T) \Or not\_at(P,T) \derives person(P), table(T).\\
\mbox{\% The persons sitting at a table cannot exceed the number of chairs there.}\\
\derives table(T), nChairs(C),\naf \countagg\{ P: at(P,T) \} \leq C.\\
\mbox{\% A person is to be seated at precisely one table.}\\
\derives person(P), \naf \countagg\{T : at(P,T)\} = 1.\\
\mbox{\% People who like each other should sit at the same table...}\\
\derives like(P1,P2), at(P1,T), \naf at(P2,T).\\
\mbox{\% ...while people who dislike each other should not.}\\
\derives dislike(P1,P2), at(P1,T), at(P2,T).\\
\end{dlvcodelong}

This encoding does not make as massive a use of aggregates as
Team Building, but it is useful to highlight a readability issue,
which also has impact on efficiency, as discussed in
Section~\ref{sec:experiments}: The last aggregate atom above could be
replaced by 
\begin{dlvcodelong}{l}
\mbox{\% A person cannot sit at two different tables...}\\
\derives person(P), at(P,T), at(P,T1), T\neq T1.\\
\mbox{\% ...and has to sit at one table at least.}\\
seated(P) \derives at(P,T).\\
\derives person(P), \naf seated(P).
\end{dlvcodelong}
This is less concise and arguably less readable.  Moreover, the
number of ground rules and constraints necessary for expressing the same
statement would grow from $k*m$ to $k*m*(m-1) + k*m + k$, where $k$
is the number of guests and $m$ the number of tables.

\paragraph{Fastfood.}
The ``Fast Food'' problem, number 662 of volume VI of the ACM
programming contests problem set archive
(\url{http://acm.uva.es/p/v6/662.html}), is specified as follows:

\begin{quote}
The fastfood chain McBurger owns several restaurants along a
highway. Recently, they have decided to build several depots along the
highway, each one located at a restaurant and supplying several of the
restaurants with the needed ingredients. Naturally, these depots
should be placed so that the average distance between a restaurant and
its assigned depot is minimized. You are to write a program that
computes the optimal positions and assignments of the depots.

To make this more precise, the management of McBurger has issued the
following specification: You will be given the positions of n
restaurants along the highway as n integers $d_1 < d_2 < \dots < d_n$
(these are the distances measured from the company's headquarter,
which happens to be at the same highway). Furthermore, a number $k$ $(k
\le n)$ will be given, the number of depots to be built.

The $k$ depots will be built at the locations of $k$ different
restaurants. Each restaurant will be assigned to the closest depot,
from which it will then receive its supplies. To minimize shipping
costs, the total distance sum, defined as

\begin{displaymath}
\sum_{i=1}^n \mid d_i - (\mbox{position of depot serving restaurant }i) \mid
\end{displaymath}

must be as small as possible.
\end{quote}

We assume that instances are given as facts of the form
$restaurant(res,d)$ representing a restaurant uniquely named $res$ at kilometer
$d$ of the highway. Moreover, a fact $nDepots(k)$ is included
which specifies $k$, the number of depots to be built.

\begin{dlvcodelong}{l}
\mbox{\% A restaurant can be a depot or not.}\\
depot(Res,D) \Or notdepot(Res,D) \derives restaurant(Res,D).\\
\mbox{\% The number of depots must be as specified.}\\
\derives nDepots(K), \naf \countagg\{Dep,D: depot(Dep,D)\} = K.\\
\mbox{\% Determine the serving depot for each restaurant.}\\
serves(Dep,Res,D) \derives restaurant(Res,ResD), depot(Dep,DepD),\\
\qquad distance(ResD,DepD,D),\\
\qquad \minagg\{Y : depot(Dep1,DepD1), distance(DepD1,ResD,Y)\} = D.\\
\mbox{\% Minimize the serving distances.}\\
\wderives serves(Dep,Res,D). [D:]\\
\mbox{\% Auxiliary predicate.}\\
distance(X,Y,D) \derives restaurant(Res1,X), restaurant(Res2,Y), X>Y,  X=Y+D.\\
distance(X,Y,D) \derives restaurant(Res1,X), restaurant(Res2,X), X<=Y, Y=X+D.
\end{dlvcodelong}

In the definition for $distance$, we have used atoms involving
built-in predicates $>$, $<=$, and $+$, which are defined on a bounded
set of integer constants. That is, these predicates define greater
than, less than or equal, and sum, respectively, on the finite set of
integers $[0,n]$. For this example domain, each instance implies an
upper bound for the integers that can occur, and we assume that the
maximum integer $n$ is chosen appropriately for each instance. Note
that atoms like $X=Y+D$ are quite different from assignment
aggregates: For the former, an admissible value range has to be
specified explicitly ($n$ for the integer range on which the predicate
is defined), while for the latter a value range is not necessary.

Note that this example involves minimization in two different ways: On
the one hand, the serving distance for a restaurant is the minimum
distance to a depot.  On the other hand, we look for a solution which
minimizes the sum of serving distances. It is important to note that
the first minimum (choosing the closest depot for every restaurant)
refers to a fixed depot assignment, whereas the second (minimizing the
sum of serving distances) is to be determined with respect to all possible
depot assignments.  It is therefore not
possible to merge the two criteria, and indeed we use different
constructs (an aggregate and a weak constraint) for representing them.

\paragraph{Input Cardinality.}

In several problems, it is important to determine the cardinality of
input relations. Doing so is simple using an assignment aggregate: If
the input predicate is $p$ and has arity $n$, we can write
\begin{dlvcodelong}{l}
cardinality\_p(C) \derives \countagg\{X1,\ldots,Xn: p(X1,\ldots,Xn)\} = C.
\end{dlvcodelong}

Note that in general this can not be achieved without assignment
aggregates as defined in Section~\ref{sec:assignmentpredicate}.
Without these, one could add some kind of domain predicate restricting
the range of the variable $C$:
\begin{dlvcodelong}{l}
cardinality\_p(C) \derives \countagg\{X1,\ldots,Xn: p(X1,\ldots,Xn)\} = C,\ domain(C).
\end{dlvcodelong}
However, since the maximum cardinality of $p$ is not known in advance,
the size of $domain$ would have to be countably infinite, which is not
feasible.

In a similar way, again by assignment aggregates,
one may compute the sum of the values of an attribute
of an input relation (e.g., compute the sum of the salaries
of the employees).

\paragraph{Fastfood Solution Checking.}

Consider a slight variation of the Fastfood problem introduced above:
Instead of computing the optimal solutions, one has to check whether a
given depot assignment is optimal and compute a witness (a depot
assignment with smaller distance sum) if it is not.  This
problem features in the First Answer Set Programming System
Competition\footnote{See
\url{http://asparagus.cs.uni-potsdam.de/contest/} .} \cite{gebs-etal-2007-lpnmr-competition}.

Here, in addition to facts $restaurant(res,d)$ (as in the Fastfood
problem input), also facts $depot(dep,d)$ will be in the input,
representing the depot assignment to be checked for
optimality.  $nDepots(k)$ is no longer part of the input.

The encoding is an elaboration of the encoding for Fastfood.  Here we
define a predicate $altdepot$, which represents an alternative depot
assignment.  Such an assignment is a witness if its distance sum is less
than the distance sum of the input depot assignment.

\begin{dlvcodelong}{l}
\mbox{\% A restaurant can be an alternative depot or not.}\\
altdepot(Res,D) \Or notaltdepot(Res,D) \derives restaurant(Res,D).\\
\mbox{\% The number of alternative depots must be equal to the number of depots.}\\
\derives \countagg\{Dep,D : depot(Dep,D)\} = N,\\
\quad \naf \countagg\{Dep,D : altdepot(Dep,D)\} = N.\\
\mbox{\% Determine the serving input depot for each restaurant.}\\
serves(Dep,Res,D) \derives restaurant(Res,ResD), depot(Dep,DepD),\\
\quad distance(ResD,DepD,D),\\
\quad \minagg\{Y : depot(Dep1,DepD1), distance(DepD1,ResD,Y)\} = D.\\
\mbox{\% Determine the serving alternative depot for each restaurant.}\\
altserves(Dep,Res,D) \derives restaurant(Res,ResD), altdepot(Dep,DepD),\\
\quad distance(ResD,DepD,D),\\
\quad \minagg\{Y : altdepot(Dep1,DepD1), distance(DepD1,ResD,Y)\} = D.\\
\mbox{\% Accept an alternative solution only if its supply costs are not greater or}\\
\mbox{\% equal than the supply costs for the input candidate.}\\
\derives \sumagg\{D,Res : serves(Dep,Res,D)\} = Cost,\\
\quad \sumagg\{D,Res : altserves(Dep,Res,D)\} \geq Cost.\\
\mbox{\% Auxiliary predicate.}\\
distance(X,Y,D) \derives restaurant(Res1,X), restaurant(Res2,Y), X>Y,  X=Y+D.\\
distance(X,Y,D) \derives restaurant(Res1,X), restaurant(Res2,X), X<=Y, Y=X+D.
\end{dlvcodelong}

It should be noted that this encoding relies heavily on assignment
aggregates. The first constraint determines the cardinality of the
input predicate $depot$ using an assignment aggregate and makes sure
that any alternative assignment has the same cardinality. The final
constraint also employs an assignment aggregate, in this case not
directly involving an input predicate, but a predicate which has a
deterministic definition ($serves$) and which involves yet another
aggregate. In fact, it is unclear if and how this constraint could be encoded
without an assignment aggregate, as the range for $Cost$ is not known
or bounded a priori.


\section{Computational Complexity of \DLPA}
\label{sec:complexity}

As for the classical non-monotonic formalisms
\cite{mare-trus-91},
two important decision problems, corresponding to two different
reasoning tasks, arise in \DLPA:

\begin{itemize}
\item[]
\textbf{Brave Reasoning:}
Given a \DLPA program \p and a ground literal $L$,
is $L$ true in some answer set of \p?

\textbf{Cautious Reasoning:} 
Given a \DLPA program \p and a ground literal $L$,
is $L$ true in all answer sets of \p?
\end{itemize}

The following theorems report on the complexity of the above reasoning
tasks for propositional (i.e., variable-free)
\DLPA programs that respect the safety restrictions imposed
in Section~\ref{sec:language}.  Importantly, it turns out that
reasoning in \DLPA does not bring an increase in computational
complexity, which remains exactly the same as for standard DLP. We
begin with programs without weak constraints,
and then discuss the complexity of full \DLPA programs
including weak constraints.

\begin{lemma}\label{lem:checking}
Deciding whether an interpretation $M$ is an answer set of a ground program
\p without weak constraints is in \CONP.
\end{lemma}

\begin{proof}
We check in \NP that $M$ is \underline{not} an answer set of \p as follows.
Guess a subset $I$ of $M$, and verify that:
(1) $M$ is not a model of \p, or
(2) $I\subset M$ and $I$ is a model of $\p^M$, the Gelfond-Lifschitz
transform of \p w.r.t.\ $M$.

The only difference w.r.t.\ the corresponding tasks
of (1) and (2) in standard \DLP is the computation of the truth valuations
of the aggregate atoms, which in turn require to compute the valuations
of aggregate functions and sets.
Computing the valuation of a ground set $T$ requires scanning
each element $\tuple{t_1,...,t_n:\mathit{Conj}}$ of $T$ and adding $t_1$
to the result multiset if $\mathit{Conj}$ is true w.r.t.\ $I$.
This is evidently polynomial, as is the application of the aggregate operators
(\countagg, \minagg, \maxagg, \sumagg, \timesagg) on a multiset.
The comparison of this result against the guards, finally, is straightforward.

Therefore, the tasks (1) and (2) are tractable as in standard \DLP.
Deciding whether $M$ is not an answer set of \p thus is in \NP;
consequently, deciding whether $M$ is an answer set
of \p is in \CONP.
\end{proof}

Based on this lemma, we can identify the computational complexity of
the main decision problems, brave and cautious reasoning.

\begin{theorem}\label{theo:brave}    
Brave Reasoning on ground \DLPA programs without weak constraints is $\SigmaP{2}$-complete.  
\end{theorem} 

\begin{proof}
We verify that a ground literal $L$ is a brave consequence of a
\DLPA program \p as follows:
Guess a set $M \subseteq \BP$ of ground atoms and check that
(1) $M$ is an answer set of \p and (2) $L$~is true w.r.t.\ $M$.
Task (2) is clearly polynomial, while (1) is in $\CONP$
by virtue of Lemma \ref{lem:checking}.
The problem therefore lies in $\SigmaP{2}$.

$\SigmaP{2}$-hardness follows from the $\SigmaP{2}$-hardness
of \DLP \cite{eite-etal-97f},
since \DLPA is a superset of \DLP.
\end{proof}

The complexity of cautious reasoning follows by similar arguments as above.

\begin{theorem}\label{theo:cautious}
Cautious Reasoning on ground \DLPA programs without weak constraints is $\PiP{2}$-complete.
\end{theorem}

\begin{proof}
We verify that a ground literal $L$ is \emph{not} a cautious consequence of 
a \DLPA program \p as follows:
Guess a set $M \subseteq \BP$ of ground atoms and check that
(1) $M$ is an answer set of \p and (2) $L$ is not true w.r.t.\ $M$.
Task (2) is clearly polynomial, while (1) is in $\CONP$,
by virtue of Lemma \ref{lem:checking}.
Therefore, the complement of cautious reasoning is in $\SigmaP{2}$,
and cautious reasoning is in $\PiP{2}$.    

$\PiP{2}$-hardness again follows from
\cite{eite-gott-95}, since \DLPA is a superset of \DLP.
\end{proof}

From these results we can derive the results for \DLPA with weak
constraints.
\begin{theorem}
For a ground \DLPA program $\p$, deciding whether an interpretation $M$ is an
answer set is $\PiP{2}$-complete, while brave and cautious reasoning
are both $\DeltaP{3}$-complete.
\end{theorem}
\begin{proof}
The key to this proof is that one can rewrite each \DLPA program \p to
another \DLPA program $\mathcal{W}(\p)$ in which no aggregates occur in weak
constraints, by replacing each aggregate literal that occurs in a weak
constraint by a new standard atom, and adding a rule with the
aggregate literal in the body and the new atom in the head.

Hardness for the $\PiP{2}$ result follows directly from item (3) of
Theorem 4.14 in \cite{leon-etal-2002-dlv}. For membership, we show
that deciding whether an interpretation $M$ is not an answer set is
$\SigmaP{2}$.  We consider $\mathcal{W}(\p)$ and $M'$, which is
obtained from $M$ by adding those new atoms that replaced aggregate
literals that are true w.r.t.\ $M$. We first test whether $M'$ is an
answer set of $\Rules{\mathcal{W}(\p)}$, which is in $\CONP$ by
Lemma~\ref{lem:checking}. If $M'$ is not an answer set, we stop and
return yes. Otherwise we determine the cost $c$ of $M'$ in polynomial
time, and guess an $M'' \subseteq \BP$. We check that $M''$ is an
answer set of $\Rules{\mathcal{W}(\p)}$ by a single call to an $\NP$
oracle, and check that the cost of $M''$ is less than $c$ in polynomial time.

For the $\DeltaP{3}$ results, hardness is an immediate consequence of
Theorem 4.8 in \cite{leon-etal-2002-dlv}. Membership can be shown
exactly as in the proof of Theorem 4.8 in \cite{leon-etal-2002-dlv},
using $\mathcal{W}(\p)$ and the fact that the necessary oracle for
determining whether an interpretation is an answer set of $\p$, the
cost of which is less than a fixed bound, is $\SigmaP{2}$ also in this
case, as argued above.
\end{proof}

The above theorems confirm that our addition of aggregates
to disjunctive logic programming does not cause any increase
in the computational complexity of the language,
and the same holds even if weak constraints are allowed.

We end this section by discussing the complexity of non-ground
programs. The problems with respect to data-complexity for \DLPA
programs (i.e.\ a program $\p$ is fixed, while the input consists of a
set of facts) have the same complexity as for propositional programs.
Concerning program
complexity (i.e.\ a program $\p$ is given as input), complexity rises
in a similar manner as for aggregate-free programs.
A non-ground program $\p$ can be reduced, by naive instantiation, to a ground
instance of the problem, the size of which is single exponential in
the size of $\p$. Informally, the complexity results thus
increase accordingly by one exponential, from 
$\CONP$ to co-NEXPTIME, $\SigmaP{2}$ to NEXPTIME$^{\NP}$, $\PiP{2}$ to co-NEXPTIME$^{\NP}$, and 
$\DeltaP{3}$ to EXPTIME$^{\SigmaP{2}}$.
These results can be derived using complexity upgrading techniques as
presented in \cite{eite-etal-97f,gott-etal-99apal}.

\section{Implementation Issues}
\label{sec:implementation}

In this section we illustrate the design of the implementation of
aggregates in the \dlv\ system.
We first briefly describe the overall architecture of \dlv,
and we then discuss the impact of the implementation of aggregates in
the system.

\begin{figure}
\begin{center}
\includegraphics[height=7.5cm]{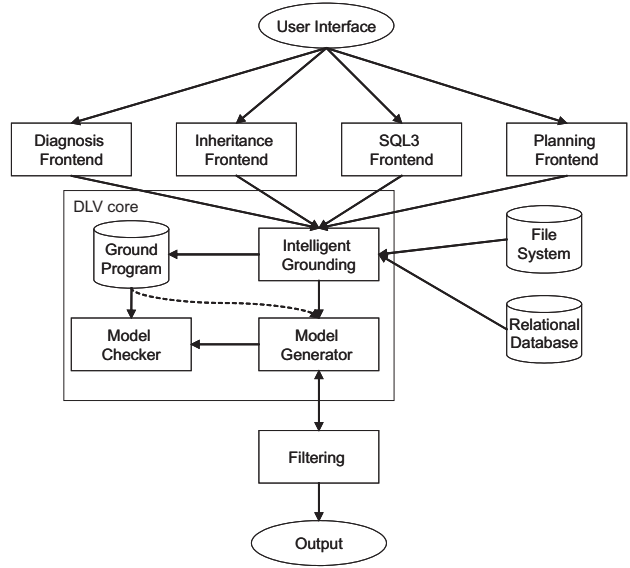}
\end{center}
\caption{The System Architecture of \dlv}
\label{fig:architecture}
\end{figure}

\subsection{\dlv\ Architecture}
\label{sec:architecture}

An outline of the general architecture of the \dlv\ system is depicted in
Figure \ref{fig:architecture}.
It includes four front-ends for solving domain-oriented
tasks; these are implemented on top of the \dlv core
by means of suitable rewriting techniques to DLP.
Clearly, the implementation of aggregates does not affect these front-ends,
even if the availability of the aggregates will allow to enhance
the front-ends and improve the expressiveness of their languages.

Instead, the implementation of aggregates heavily affects the \dlv
core, which we describe next.
Input data can be supplied by regular files, and also by relational databases.
The \dlv\ core then produces answer sets one at a time, and each time an
answer set is found, the ``Filtering'' module is invoked,
which performs post-processing
(dependent on the active front-ends) and controls continuation or
abortion of the computation.

The \dlv\ core consists of three major components: the ``Intelligent
Grounding'', the ``Model Generator'', and the ``Model Checker'' modules that share
a principal data structure, the ``Ground Program''. The ``Ground Program'' is created by
the ``Intelligent Grounding'' using differential (and other advanced) database techniques together
with suitable data structures, and used by the ``Model Generator'' and
the ``Model Checker''. The Ground Program is guaranteed to have exactly the same answer sets as the original program.
For some syntactically restricted classes of programs (e.g.\ stratified programs), the ``Intelligent Grounding'' module already computes
the corresponding answer sets.

For harder problems, most of the computation is performed by the ``Model
Generator'' and the ``Model Checker''.  Roughly, the former
produces some candidate answer sets (models) \cite{fabe-etal-99b,fabe-etal-2001a}, the stability
and minimality of which are subsequently verified by the latter.

The ``Model Checker'' (MC) verifies whether the model at hand is
an answer set.  This task is very hard in general, because
checking the stability of a model is known to be co-\NP-complete.
However, MC exploits the fact that minimal model checking --- the
hardest part --- can be efficiently performed for the relevant
class of \emph{head-cycle-free} (HCF) programs
\cite{bene-dech-94,leon-etal-97b}.

\subsection{Implementation of Aggregates in \dlv}
Implementing aggregates in the \dlv\ system, has had a strong impact
on \dlv\ requiring many changes to the modules of the \dlv\ core,
and, especially, to the ``Intelligent Grounding'' (IG)
and to the ``Model Generator'' (MG) modules.
We next describe the main changes carried out in the modules of \dlv core
to implement aggregates.

\subsubsection{Intelligent Grounding}

The changes performed in the Intelligent Grounding module
to implement aggregates in \dlv
can be summarized in three main activities:
Standardization, Instantiation Procedure
(the main task), and Duplicate Sets Recognition.

\paragraph{Standardization.}
After parsing, each aggregate $A$ is transformed such that both guards are
present and both $\prec_1$ and $\prec_2$ are set to $\leq$.
The conjunction $\mathit{Conj}$ of the symbolic set of $A$ is replaced by
a single, new atom $Aux$ and a rule $Aux\derives \mathit{Conj}$ is added
to the program
(the arguments of $Aux$ being the distinct variables of $\mathit{Conj}$).

\paragraph{Instantiation Procedure.}
The goal of the instantiator is to generate a ground program that has
precisely the same answer sets as the theoretical instantiation $\GP$,
but is as small as possible.  The instantiation of standard \dlv
proceeds bottom-up following the dependencies induced by the rules,
and, in particular, respecting the ordering imposed by
negation-stratification where this is possible.
\dlv's instantiator produces only those instances of a predicate
which can potentially become true
\cite{fabe-etal-99c,leon-etal-2001a}, where a ground atom $A$ can
potentially become true only if we have generated or may generate a
ground instance of a rule with $A$ in the head. Ground atoms, which
have determined to be true or false in any answer set, are instead
partially evaluated, that is if a literal it occurs in is true, that
literal is omitted from the ground rule to be generated; if that
literal is false, the ground rule it would occur in will not be
generated.

For programs containing stratified aggregates strategy is extended
such that the order of processing respects aggregate
stratification. In this way, any truth-values (true, false or
potentially true) of nested atoms, which can be determined during
grounding, have already been determined before the aggregate atom
itself is instantiated.

When processing a rule containing an aggregate atom we proceed as
follows.  Assume that ``$H \derives B, aggr.$'' is the rule $r$ which is to be
processed, where $H$ is the head of the rule, $B$ is the conjunction
of the standard body literals in $r$, and $aggr$ is a standardized
aggregate literal over a symbolic set
\mbox{$\{\mathit{Vars}\!\!:\!\!Aux\}$}.  First we compute an
instantiation $\overline{B}$ for the literals in $B$; this also binds
the global variables appearing in $Aux$.  The (partially bound) atom
$\overline{Aux}$ is then matched against its extension (which is
already available as the computation follows aggregate-stratification
as discussed above),
all matching facts are collected, and a set of pairs
\[
\{ \tuple{\theta_1(\mathit{Vars})\!:\!\theta_1(\overline{Aux})},...,\tuple{\theta_n(\mathit{Vars})\!:\!\theta_n(\overline{Aux})} \}
\]
is generated, where $\theta_i$ is a substitution for the local
variables in $\overline{Aux}$ such that $\theta_i(\overline{Aux})$ is
a potentially true instance of $\overline{Aux}$. For all
$\sigma(\overline{Aux})$ which are true or false instances of
$\overline{Aux}$, the aggregate is partially evaluated, which is done
by methods that depend on the aggregate function involved.  Note that
in this way aggregates will only ground atoms the truth-value of which
can not be determined already by the instantiator.

Note that for several classes of programs, the instantiator guarantees
complete evaluation. If a predicate is defined by a subprogram of such
a class, no ground atom of it will be generated. In particular, if the
predicate $Aux$ of a standardized aggregate is defined by such a
program, the aggregate function can be fully evaluated by the
instantiator. One notable class, for which this is possible, are
non-disjunctive negation-stratified programs. Therefore, by the
definition of assignment aggregates, the value of the aggregate
function inside an assignment aggregate can always be determined by
the instantiator, thus providing a binding for the assigned variable
(or no binding if the function evaluates to $\bot$). An assignment
aggregate thus is treated like a unary positive atom which has at most
one true matching instance.

If a non-assignment aggregate literal can be fully evaluated by the
instantiator, its truth-value will be determined by computing the
value of the aggregate function and comparing it to the guards. If it
evaluates to true, it is removed from the ground rule, if it evaluates
to false, the ground rule is simply discarded, thus partially
evaluating the ground rule an the aggregate literal.

The same process is then repeated for all further instantiations of the
literals in $B$.

\begin{example}
Consider the rule $r$:
\vspace*{1.5mm}
\begin{dlvcode}
p(X) \derives q(X), 1 < \countagg\{Y:a(X,Y), \naf b(Y)\}.
\vspace*{1mm}
\end{dlvcode}
The standardization rewrites $r$ to:
\vspace*{1.5mm}
\begin{dlvcode}
p(X) \derives q(X), 2 \leq \countagg\{Y:aux(X,Y)\}\leq \infty. \\
aux(X,Y)\derives a(X,Y), \naf b(Y).
\vspace*{1mm}
\end{dlvcode}
Suppose that the instantiation of the rule for $aux$ generates 3 potentially
true facts $aux(1,a)$, $aux(1,b)$, and $aux(2,c)$.
If the potentially true facts for $q$ are $q(1)$ and $q(2)$,
the following ground instances are generated:
\vspace*{1mm}
\begin{dlvcode}
p(1) \derives q(1), 2\! \leq\! \countagg\{\tuple{a\!:\!aux(1,a)},
\tuple{b\!:\!aux(1,b)}\}\!\leq\! \infty. \\
p(2) \derives q(2), 2 \leq \countagg\{ \tuple{c : aux(2,c)} \}\leq \infty.
\vspace*{1mm}
\end{dlvcode}
Note that a ground set contains only those $aux$ atoms which are
potentially true.
\punto\end{example}

\paragraph{Duplicate Sets Recognition.}
To optimize the evaluation during instantiation and especially afterward,
we have designed a hashing technique which
recognizes multiple occurrences of the same set in the program,
even in different rules, and stores them only once.
This saves memory (sets may be very large), and
implies a significant performance gain,
especially during model generation where sets are frequently
manipulated by the backtracking process.
\begin{example}
Consider the following two constraints:
\begin{dlvcode}
c_1\!:\ \  \derives\ \ 10 \leq \maxagg\{V : d(V,X)\}. \\
c_2\!:\ \  \derives\ \ \minagg \{Y : d(Y,Z)\} \leq 5.
\end{dlvcode}
Our technique recognizes that the two sets are equal,
and generates only one instance which is shared by $c_1$ and $c_2$.

To see the impact of this technique, consider a situation in which the two constraints additionally
contain another standard literal $p(T)$:
\begin{dlvcode}
c_3\!:\ \  \derives\ \ p(T), 10 \leq \maxagg\{V : d(V,X)\}. \\
c_4\!:\ \  \derives\ \ p(T), \minagg \{Y : d(Y,Z)\} \leq 5.
\end{dlvcode}
Here, $c_3$ and $c_4$ have $n$ instances each, where $n$ is
the number of potentially true atoms matching $p(T)$.  By means of our technique, all instances of the constraint atoms in $c_3$ and $c_4$ share one common set, 
reducing the number of instantiated sets from $2*n$ to $1$.
\punto\end{example}

\subsubsection{Model Generator}

In our implementation, an aggregate atom will be assigned a truth-value
just like a standard atom. However, different from a standard atom,
its truth-value also depends on the valuation of the aggregate
function and thus on the truth-value of the nested
predicates. Therefore, an aggregate atom adds an implicit constraint
on models and answer sets: The truth-value assigned to the aggregate
atom must correspond to the truth-value obtained by the valuation.

We have designed an extension of the Deterministic Consequences
operator of the \dlv system \cite{fabe-etal-99b,fabe-2002} for \DLPA
programs which accounts for these additional implicit
constraints. As for rules, we differentiate between ``forward
propagation'' (when an aggregate atom is assigned a truth-value because
of the valuation of its aggregate function) and ``backward
propagation'' (when a nested atom is derived in order to make the
valuation of the aggregate atom compliant with its assigned
truth-value). 

We have extended the Dowling and Gallier algorithm
\cite{dowl-gall-84} (in the version of \cite{mino-88}) to deal with aggregates,
and we compute the fixpoint of the enhanced Deterministic Consequences
operator in linear time. To achieve this, we have
endowed aggregate atoms with datastructures similar to those used in
rules. In particular, all aggregate atoms have a lower and upper bound
holding the minimum and maximum value of the aggregate function w.r.t.\ the
interpretation at hand to
efficiently determine whenever an aggregate atom becomes true or
false. \minagg and \maxagg hold additional values for differentiating
between undefined and true nested atoms. Moreover, for each standard
atom we keep an index of aggregate sets in which it occurs to update
these counters in an efficient way.

Forward propagation can then be achieved comparatively easily: whenever
a standard atom is assigned a truth-value (other than undefined), the
bounds and additional data of all aggregate functions it occurs in are
updated. Where the bound range is fully covered by the guard range, the
aggregate atom is derived as true. If the bound range and the
guard range do not intersect, it is derived as false. For backward
propagation, whenever an aggregate atom gets a truth-value other than
undefined or a non-undefined aggregate atom has an update of its
bounds, several checks for inferences are performed, dependent on the
type of aggregate function. For example, if there exists a tuple
$\tuple{k,\ldots\ :\!a}$ in a ground \sumagg aggregate which is true,
such that $a$ is undefined and the lower bound plus $k$ is greater than the upper guard,
then $a$ can be derived as false. In order to make these checks
efficient, the set of entries in the ground aggregate set is stored in
a structure which is ordered on the projected term.

\begin{example}
Let us consider some of the propagations that are done for the following ground program.
\begin{dlvcode}
a(1) \Or b(1).\qquad a(2) \Or b(2).\\
\derives \sumagg\{ \tuple{1\!:\!a(1)}, \tuple{2\!:\!a(2)} \} < 3.\\
cs \derives \countagg\{ \tuple{1\!:\!a(1)}, \tuple{2\!:\!a(2)} \} \geq 2.\\
c(1) \derives cs.\qquad c(2) \Or c(3) \derives cs.\qquad\derives c(1), d(1).\\
d(2) \derives \minagg\{ \tuple{1\!:\!c(1)}, \tuple{2\!:\!c(2)}, \tuple{3\!:\!c(3)} \} < 2.\\
d(1) \derives \maxagg\{ \tuple{1\!:\!c(1)}, \tuple{2\!:\!c(2)}, \tuple{3\!:\!c(3)} \} \geq 3.
\end{dlvcode}

At the very beginning, the internal datastructures of the aggregate
atoms are initialized. $\sumagg\{ \tuple{1\!:\!a(1)},
\tuple{2\!:\!a(2)} \} < 3$ gets bounds $[0,3]$ and guards $[0,2]$ (the
guards are normalized to be inclusive). In a similar way, $\countagg\{
\tuple{1\!:\!a(1)}, \tuple{2\!:\!a(2)} \} \geq 2$ gets bounds $[0,2]$
and guards $[2,\infty]$. $\minagg\{ \tuple{1\!:\!c(1)},
\tuple{2\!:\!c(2)}, \tuple{3\!:\!c(3)} \} < 2$ is initialized with
bounds $[-\infty,+\infty]$ (because the value of the aggregate
function may become undefined) and guards $[0,1]$, and in
addition $\mathtt{minTrue}=\infty$ and $\mathtt{minUndef}=1$ for
keeping track of possible minima. In a similar way,
$\maxagg\{\tuple{1\!:\!c(1)}, \tuple{2\!:\!c(2)}, \tuple{3\!:\!c(3)}
\} \geq 3$ is initialized with bounds $[-\infty,+\infty]$, guards
$[3,\infty]$, and special values $\mathtt{maxTrue}=-\infty$ and
$\mathtt{maxUndef}=3$.

In the first step, $\sumagg\{ \tuple{1\!:\!a(1)}, \tuple{2\!:\!a(2)}
\} < 3$ is derived false in order to satisfy the first constraint. In order
to look for possibilities for backward propagation, the elements of
the multiset are examined in a descending order, beginning with the
largest one. For each element, we check whether the bound minus the element value is less than or equal to the upper guard, as the condition of these elements
must become true. So for $\tuple{2\!:\!a(2)}$, we obtain $3-2\leq 2$
and we make a derivation establishing the fact that $a(2)$ must be
true. In a similar manner, we obtain that $a(1)$ must be true. Since
both $a(1)$ and $a(2)$ each occur in a single rule head, they are
derived as definitely true, being supported by the respective rule,
which in turn causes $b(1)$ and $b(2)$ to be derived as false. Moreover, the
truth of $a(2)$ causes the bounds of $\countagg\{ \tuple{1\!:\!a(1)},
\tuple{2\!:\!a(2)} \} \geq 2$ to become $[1,2]$, which due to the
truth of $a(1)$ then become $[2,2]$, causing the aggregate atom to
become true.

As a consequence, also $cs$ and $c(1)$ become true, while $c(2)$ and
$c(3)$ remain undefined. So in $\minagg\{ \tuple{1\!:\!c(1)},
\tuple{2\!:\!c(2)}, \tuple{3\!:\!c(3)} \} < 2$, $\mathtt{minTrue}$
becomes $1$, while $\mathtt{minUndef}$ becomes $2$, so its bounds
become $[1,1]$, and the aggregate atom becomes true, causing also
$d(2)$ to become true. For $\maxagg\{ \tuple{1\!:\!c(1)},
\tuple{2\!:\!c(2)}, \tuple{3\!:\!c(3)} \} \geq 3$, $\mathtt{maxTrue}$
becomes $1$ and $\mathtt{maxUndef}$ becomes $3$, causing the bounds to
become $[1,3]$. 

Moreover, $d(1)$ becomes false because of the
constraint $\derives c(1),d(1).$ Therefore, the aggregate atom $\maxagg\{ \tuple{1\!:\!c(1)},
\tuple{2\!:\!c(2)}, \tuple{3\!:\!c(3)} \} \geq 3$ is derived as false.
We then examine the elements of the multiset, starting with the
greatest. If a condition of the element is undefined and its value is
between the guards (inclusively), that condition must become false.
In our example, for $\tuple{3\!:\!c(3)}$ this holds (the guards are
$[3,\infty]$) and so we derive the falsity of $c(3)$.  For
$\tuple{2\!:\!c(2)}$ we cannot do this, as $2$ is not within the
guards. Indeed, $c(2)$ is eventually derived true in order to satisfy
the rule $c(2) \Or c(3) \derives cs$.

In this example, the Deterministic Consequence operator has thus
already determined the answer set $\{a(1), a(2), cs, c(1), c(2),
d(2)\}$, as no undefined atoms are left.
\end{example}

\subsubsection{Model Checker}

The stratification constraint that we have imposed on \DLPA aggregates,
allows us to treat aggregate literals as negative literals in the reduct
(see Section \ref{sec:semantics}),
and minimize the impact of aggregates on answer set checking.

The Model Checker (MC) receives a model $M$ in input,
and checks whether $M$ is an answer set of the instantiated program $\p$
(see Subsection \ref{sec:architecture}).
To this end, it first computes the reduct $\p^M$, by
(i) deleting the rules having a false aggregate literal or a false negative
literals (w.r.t.\ $M$) in their bodies, and
(ii) removing the aggregates literals and the negative literals
from the bodies of the remaining rules.
Since the resulting program is aggregate-free,
the standard \dlv techniques can then be applied
to check whether $\p^M$ is an answer set.
Thus, no further change is needed in MC, after the modification
of the procedure computing the reduct.

\section{Experiments and Benchmarks}
\label{sec:experiments}

\subsection{Compared Methods, Problems and Data}

To assess the usefulness of the proposed \DLP language extension
and evaluate its implementation,
we compare the following two methods on some relevant benchmark
problems: 

\medskip

\begin{tabular}{@{}l@{\ }p{10.5cm}}
\dlva & Encode each problem in \DLPA and solve it using our extension
of \dlv with aggregates.
\\
\dlv & Encode the problem in standard \DLP and solve it
using standard \dlv.

To generate \DLP encodings from \DLPA encodings,
suitable logic definitions of the aggregate functions are employed
(which are recursive for $\countagg$, $\sumagg$, and $\timesagg$).
\end{tabular}

\medskip

We compare these methods on three benchmark problems: Time Tabling,
Seating, and Fastfood.
\textbf{Time Tabling} is a classical scheduling problem.
In particular, we consider the problem of scheduling the timetable of
lectures which some groups of students have to take using a
number of real-world instances from the University of Calabria where
instance $k$ deals with $k$ groups of students.

\textbf{Seating} is the problem described in Section~\ref{sec:KR}.
We consider four (for small instances with at most four tables) or
five (for larger instances with at least five tables)
seats per table, with increasing numbers of tables and persons
(and $\mathit{numPersons} = \mathit{numSeats} * \mathit{numTables}$).

For each problem size (i.e., seats per tables/tables configuration),
we consider classes with different numbers of like and dislike 
constraints, where the percentages are relative to the maximum numbers of
like and dislike constraints, respectively, such that the problem is not
over-constrained.%
\footnote{Beyond these maxima there is trivially no solution.}
In particular, we consider the following classes:
\begin{itemize}
\item no like/dislike constraints at all;
\item 25\% like constraints;
\item 25\% like and 25\% dislike constraints;
\item 50\% like constraints;
\item 50\% like and 50\% dislike constraints.
\end{itemize}
For each problem size, we randomly generated 10 instances for each of these
classes,
50 instances for each problem size overall.

We use the \DLPA encoding reported in Section~\ref{sec:KR}.
All encodings and benchmark data are available on the web
at \url{http://www.dlvsystem.com/examples/} in the files
{aggregates-timetabling.zip},
{aggregates-seating.zip}, and
{aggregates-fastfood.zip}.

\textbf{Fastfood} is the problem described in Section~\ref{sec:KR}.
The concrete instances consist of service station data of the
company ``Tank\&Rast'' which runs the majority of service stations on
German motorways. This data has been obtained from the company website
\url{http://www.rast.de/standorte/}. 

The instances are grouped by motorway and vary over the
number of depots to be built, ranging from 0 to the total number of
restaurants along the motorway. The maximum length of any motorway is
910 kilometers, the maximum number of restaurants per motorway is 49.

\subsection{Results and Discussion}

We ran all benchmarks on an Intel dual Xeon 3GHz machine, using Debian
GNU/Linux sarge with kernel version 2.4.27 and \dlv release
2006-07-14. We allowed a maximum running time of 1800 seconds per
instance and a maximum memory usage of 256MB.

\begin{table}
\caption{Experimental Results for Timetabling}
\label{tab:timetabling}

\begin{tabular}{c||rr|rr}
Number of  & \multicolumn{2}{c|}{Execution Time} & \multicolumn{2}{c}{Instantiation Size} \\
Groups & \dlv & \dlva & \dlv & \dlva \\
\hline
1 & 3.45  & 0.22  & 91337 & 7092 \\
2 & 12.40  & 0.77  & 178756 & 14209 \\
3 & 32.63  & 1.57  & 265250 & 21200 \\
4 & 59.49  & 2.73  & 367362 & 29377 \\
5 & 90.93  & 4.18  & 437018 & 36517 \\
6 & 129.44  & 5.76  & 519568 & 43385 \\
7 & 153.30  & 7.98  & 607099 & 50731 \\
8 & 216.12  & 11.70  & 762026 & 62513 \\
9 & -  & 16.51  & 944396 & 74772
\end{tabular}
\end{table}

Cumulated results for Timetabling and Seating are provided in
Tables~\ref{tab:timetabling} and \ref{tab:seating}, respectively.
For Timetabling we report the execution time and the
size of the residual ground instantiation (the total number of atoms
occurring in the instantiation, where multiple occurrences of the same
atom are counted separately and atoms occurring in the sets of
the aggregates are considered, too).
For Seating, the execution time is the average running time
over all instances of the same size.

A ``-'' symbol in the tables indicates that the corresponding instance
(some of the instances of that size, for Seating)
was not solved within the allowed time and memory limits.

On both problems, \dlva clearly outperforms \dlv.
On Timetabling, the execution time of \dlva is one order of magnitude
lower than that of \dlv on all problem instances, and \dlv could not solve
the last instances within the allowed memory and time limits.
On Seating, the difference became even more significant.
\dlv could solve only instances of small size (up to 16 persons
-- 4 tables, 4 seats per table), while \dlva could solve significantly larger
instances in reasonable time.

The data on the instantiation sizes provides an explanation
for the large difference between the execution times of 
\dlv and \dlva.
Thanks to aggregates, the \DLPA encodings of Timetabling and Seating
are far more succinct than the corresponding encodings in standard
\DLP. This also reflects in the ground instantiations
of the programs.
Since the evaluation algorithms are exponential in the size of
the instantiation (in the worst case), the execution times of \dlva
turn out to be much shorter than those of \dlv.

\begin{table}
\caption{Experimental Results for Seating}
\label{tab:seating}

\begin{tabular}{c||rr|rr}
Number of  & \multicolumn{2}{c|}{Exec.\ Time} &
\multicolumn{2}{c}{Instantiation Size} \\
 Persons   &   \dlv &    \dlva & \dlv &   \dlva \\
\hline
8 & 0.01 & 0.01 & 228 & 72\\
12 & 0.0155 & 0.01 & 710 & 176\\
16 & 10.294 & 0.01 & 1621 & 348\\
25 & - & 0.01 & 4744 & 960\\
50 & - & 0.0505 & 35779 & 5443\\
75 & - & 0.1869 & 118167 & 15744\\
100 & - & 0.5371 & 277035 & 34221\\
125 & - & 1.2619 & 537635 & 63358\\
150 & - & 2.6204 & 925055 & 105476\\
175 & - & 4.854 & 1464260 & 162773
\end{tabular}
\end{table}

For Fastfood, we report only on motorways yielding hard instances in
Figure~\ref{fig:fastfood}. We have omitted the graph for A7, as it is
very similar to that of A3. In each graph, the horizontal axis
represents the number of depots to be built, while the vertical axis
stands for execution time. For all motorways, we can observe an
interesting easy-hard-easy pattern with increasing number of
depots. This is expected, as most possibilities for placing depots
exist when the number of depots to be built is about half of the
number of restaurants. We also observe that the average execution
times strongly depend on the number of restaurants.

It is easy to see that the encoding greatly benefits from the use of
aggregates: Whenever there are instances that cannot be solved within
the time limit, the version with aggregates manages to solve strictly
more instances without timing out. Also when looking at the amount of
time needed, the version with aggregates is always faster, and the
advantage becomes more pronounced with rising difficulty of the
instances, yielding speedups of up to $3.5$. The computational
benefits for this problem are not as dramatic as for Timetabling and
Seating, but still quite notable.

\newcommand{\mycap}[1]{\hspace*{5.5em}#1}

\begin{figure}
\caption{Results for Fastfood on German Motorways A1-A5 and A8}
\label{fig:fastfood}
\begin{tabular}{cc}

\\ & \\

\begin{minipage}{.45\linewidth}
\includegraphics[width=0.99\textwidth]{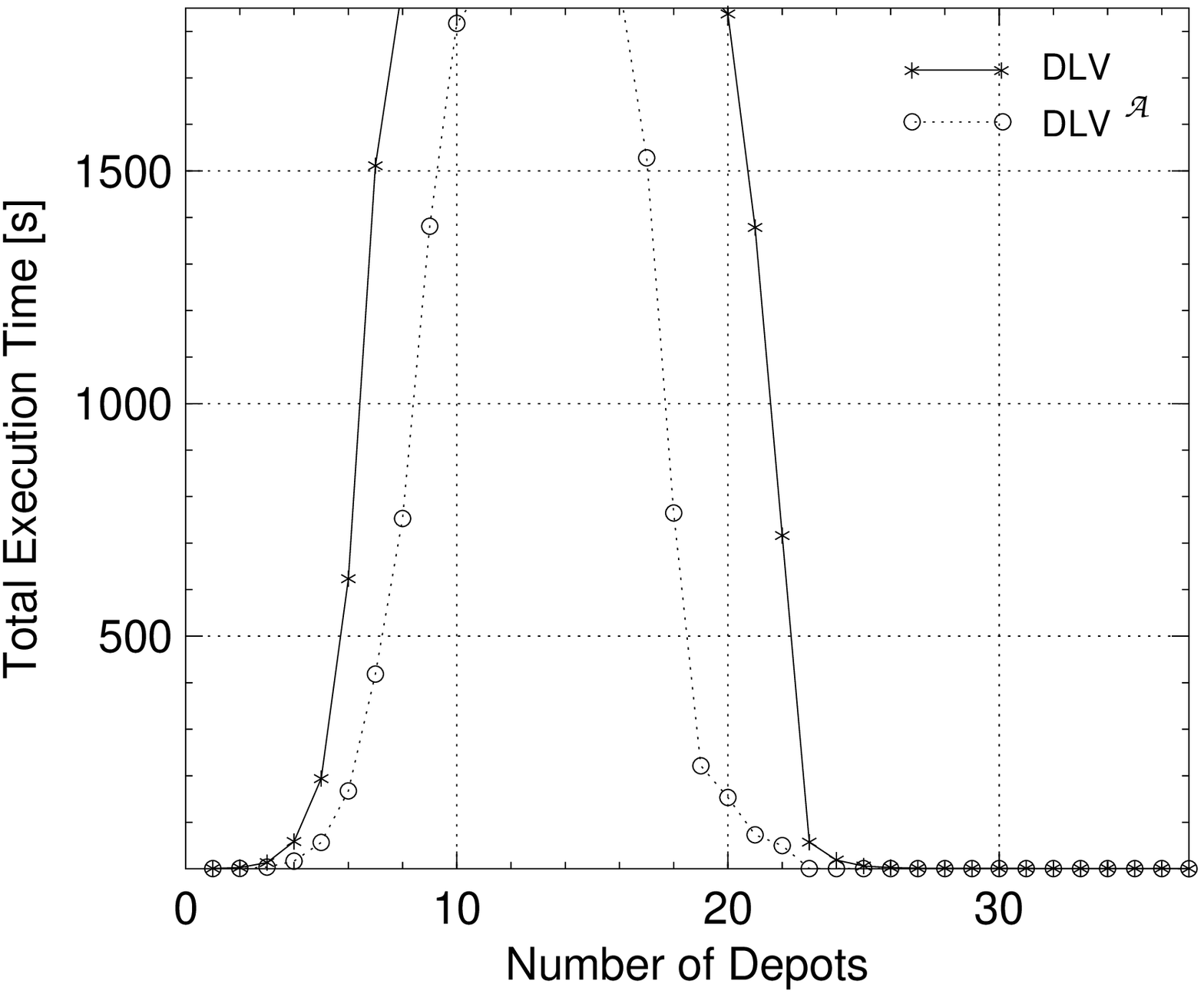}
\mycap{German Motorway A1}
\label{fig:fastfood_A1}
\end{minipage}

&

\begin{minipage}{.45\linewidth}
\includegraphics[width=0.99\textwidth]{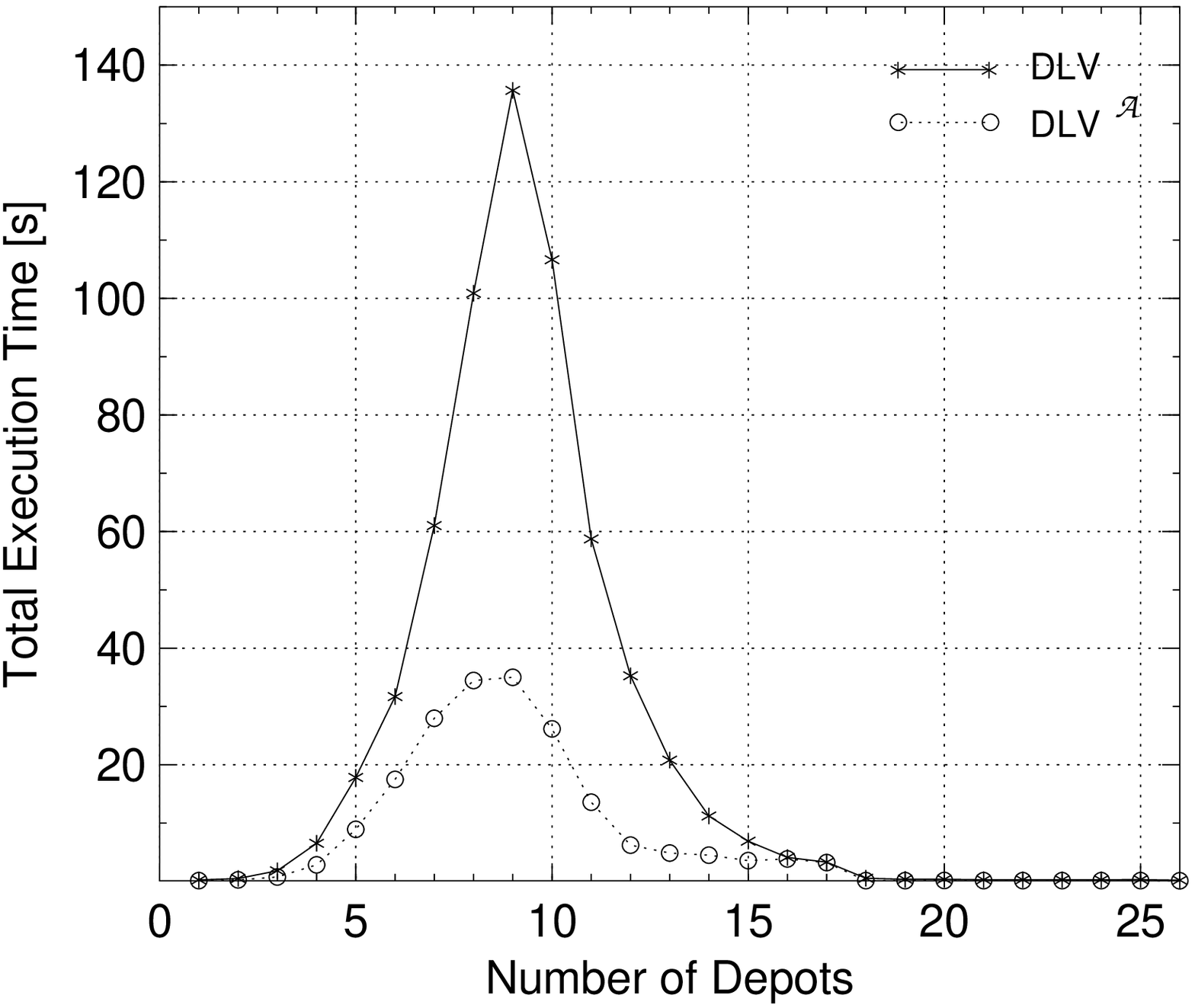}
\mycap{German Motorway A2}
\label{fig:fastfood_A2}
\end{minipage}

\\ & \\

\begin{minipage}{.45\linewidth}
\includegraphics[width=0.99\textwidth]{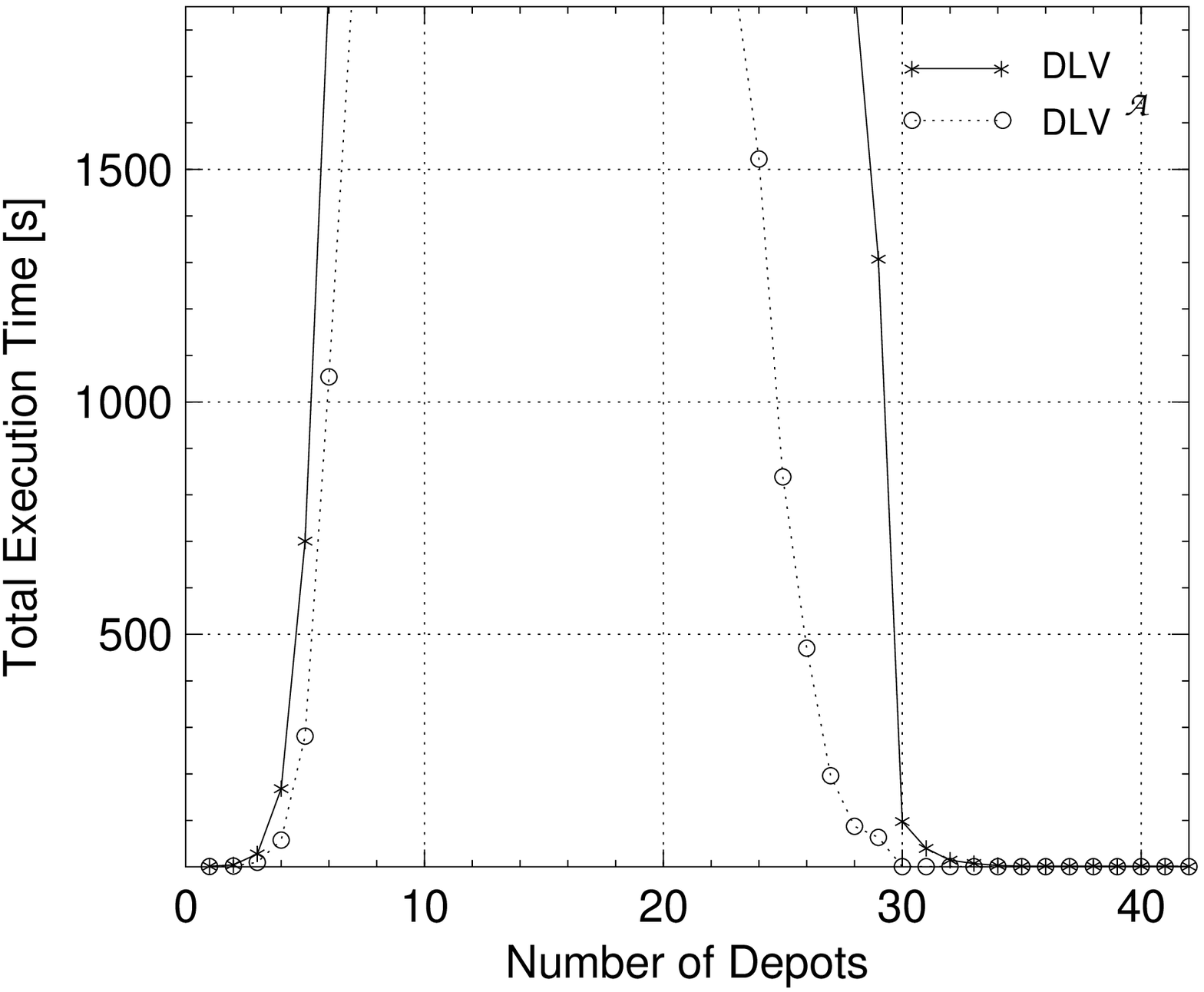}
\mycap{German Motorway A3}
\label{fig:fastfood_A3}
\end{minipage}

&

\begin{minipage}{.45\linewidth}
\includegraphics[width=0.99\textwidth]{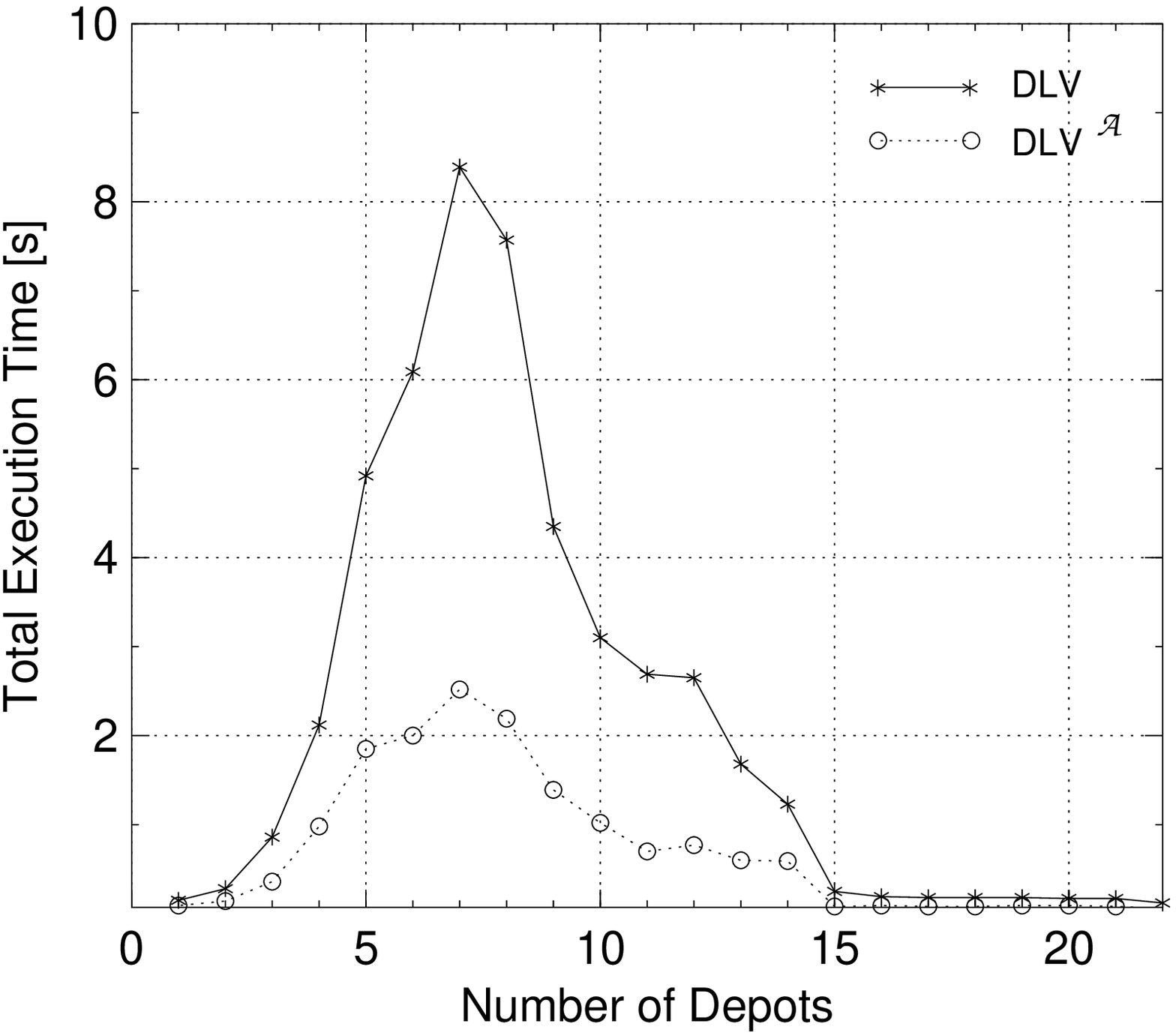}
\mycap{German Motorway A4}
\label{fig:fastfood_A4}
\end{minipage}

\\ & \\

\begin{minipage}{.45\linewidth}
\includegraphics[width=0.99\textwidth]{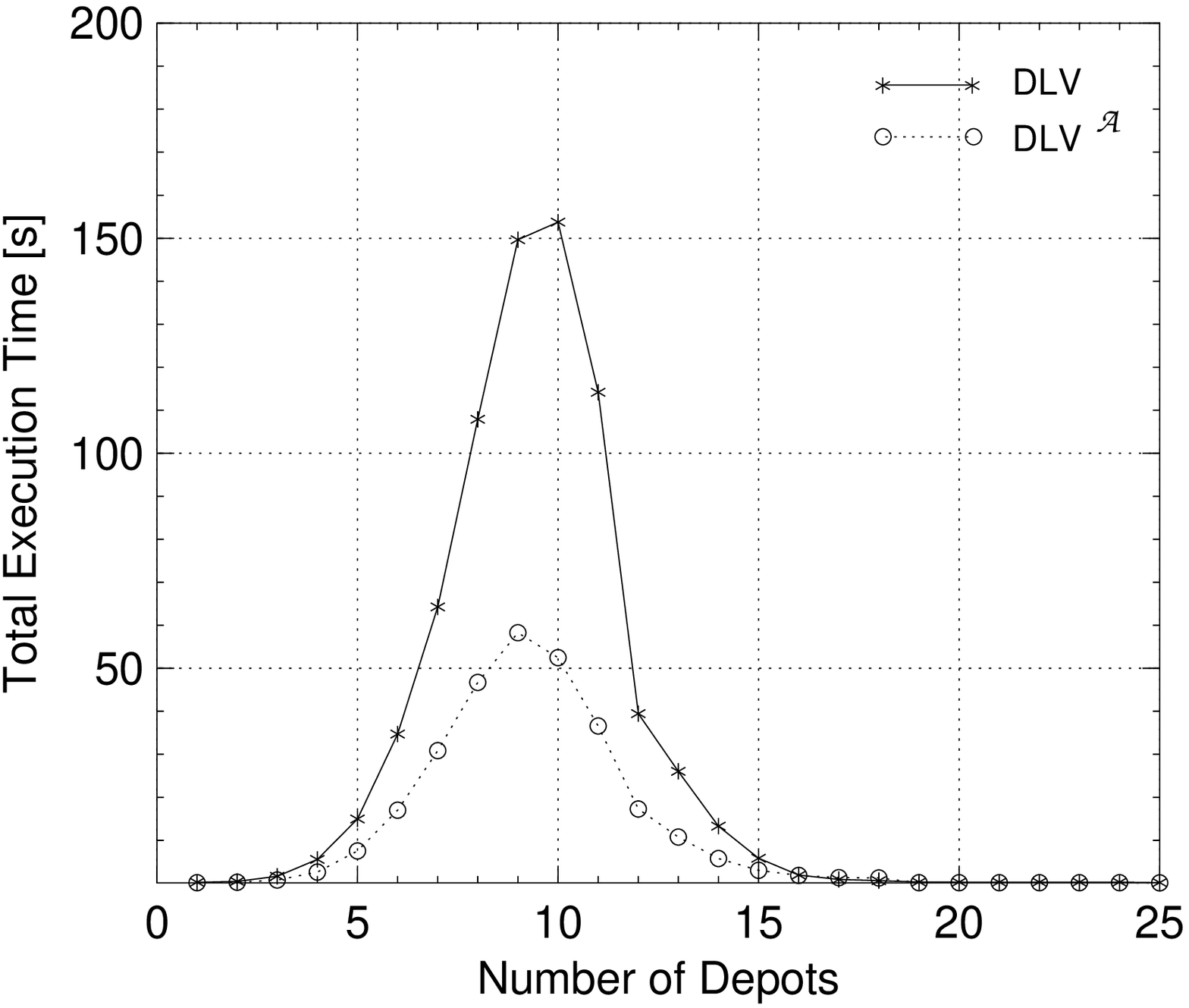}
\mycap{German Motorway A5}
\label{fig:fastfood_A5}
\end{minipage}

&

\begin{minipage}{.45\linewidth}
\includegraphics[width=0.99\textwidth]{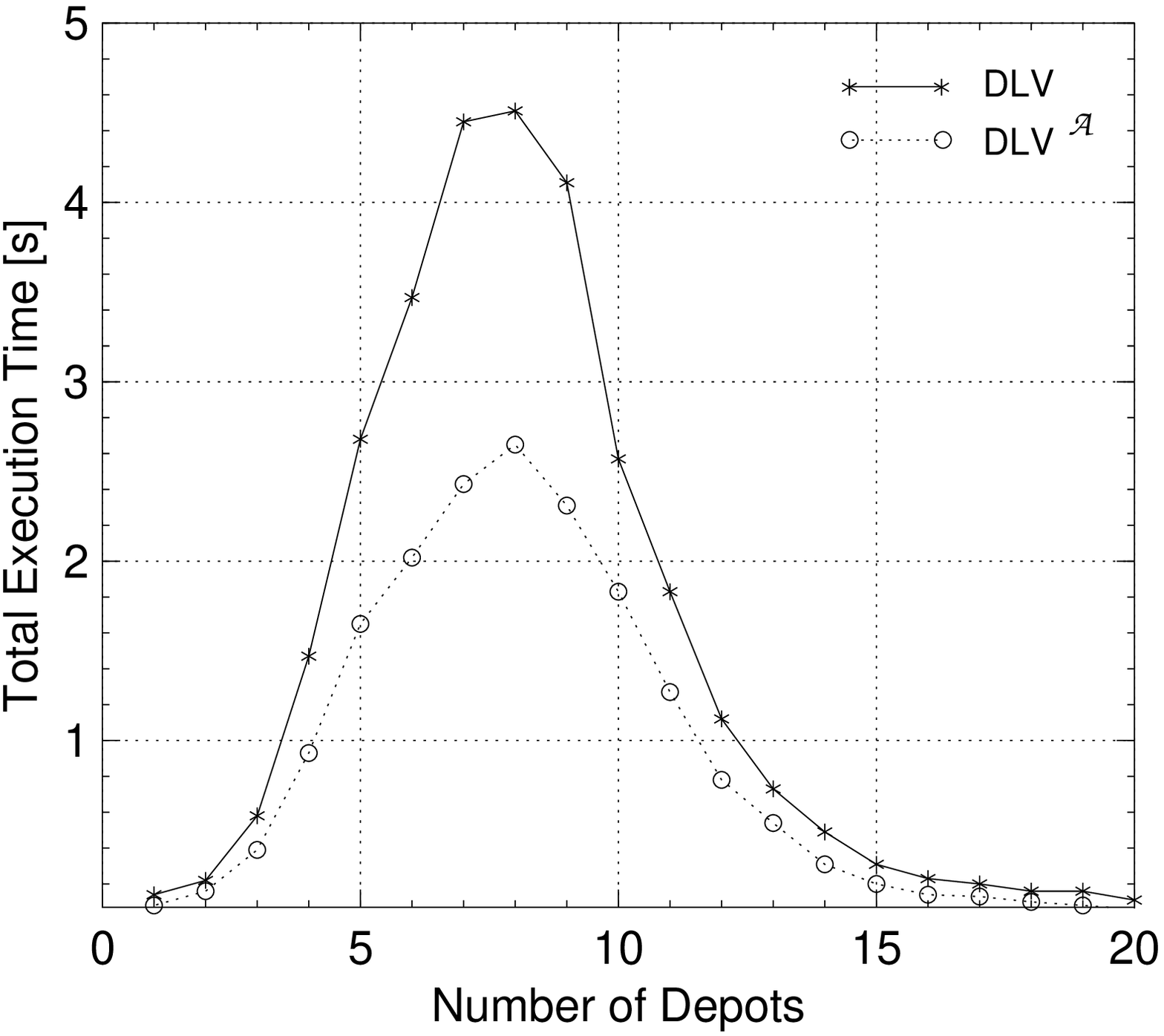}
\mycap{German Motorway A8}
\label{fig:fastfood_A8}
\end{minipage}

\end{tabular}
\end{figure}

\section{Related Work}\label{sec:related}

Aggregates have been studied fairly extensively in the context of
databases and logics for databases, see \cite{hell-etal-2001-jacm} for
a summary. The logics studied in this setting are typically
first-order logic endowed with some sort of aggregation operators,
which are used to express queries. In such logics there is no concept
of recursive definitions, and the aggregations therefore occur in a
stratified way. Moreover, as shown in \cite{hell-etal-2001-jacm}, the
expressivity of these languages suffers from similar limitations as
standard first-order logics for query answering.

Aggregate functions in logic programming languages appeared already in
the 1980s, when their need emerged in deductive databases like
LDL~\cite{chim-etal-90} and were studied in detail, cf.\
\cite{ross-sagi-1997,kemp-rama-1998}. However, the first implementation
in Answer Set Programming, in the Smodels system, has been fairly
recent \cite{simo-etal-2002}.

\subsection{Aggregate-Stratification}
\label{sec:related-stratification}

The discussion on the ``right'' semantics for aggregate-unstratified
programs is still going on in the DLP and Answer Set Programming (ASP)
communities. Several proposals
have been made in the literature, which can roughly be grouped as
follows: In \cite{eite-etal-97g,gelf-2002,dell-etal-2003b}, aggregate
atoms are basically treated like negative literals. In
\cite{niem-etal-99}, only aggregates involving cardinality and sum are
considered; as argued in \cite{ferr-lifs-2005} this semantics is not
intuitive for aggregates which are not monotonic, such as sum
aggregates involving negative summands. In
\cite{pelo-2004,pelo-etal-2004a}, a family of semantics, which
extend completion, stable and well-founded semantics, is defined
by means of operator fixpoints, approximations and
transformations; a very similar approach has been given in
\cite{son-etal-2005-tr} and \cite{son-pont-2007-tplp}. In
\cite{fabe-etal-2004-jelia}, a semantics based on a modified program
reduct has been defined, for which alternative characterizations have
been provided in
\cite{ferr-2005-lpnmr,cali-etal-2005-ijcai,fabe-2005-lpnmr}. All of these
four groups of semantics differ on certain language fragments; but
they coincide on aggregate-stratified programs. Finally, in
\cite{mare-etal-2004-lpnmr,liu-trus-2005-aaai}, semantically
restricted aggregates are considered, on which the newer proposals
coincide; but still the first group of semantics
\cite{eite-etal-97g,gelf-2002,dell-etal-2003b} differs even on these
programs. To illustrate the difficulties with unstratified aggregates, we
look at a simple example:

\begin{example}\label{ex:unstratification}
Consider the (aggregate-unstratified) program
consisting only of the rule 
\[
p(a) \derives \countagg \{ X:p(X)\}=0.
\]
As neither $\{p(a)\}$ nor $\emptyset$ is an intuitive
meaning for the program, one would expect that this program admits no
answer sets. In this case, the role of the aggregate literal is
similar to a negative literal. And indeed, approaches like
\cite{dell-etal-2003b,gelf-2002,eite-etal-97g}
treat aggregates like negative literals.

However, consider a slight modification of this program, containing
only the rule
\[
p(a) \derives \countagg \{X:p(X)\}>0.
\]
If the aggregate is treated like a negative literal, this program
allows for two answer sets $\{p(a)\}$ and $\emptyset$. Other
approaches \cite{pelo-2004,fabe-etal-2004-jelia,mare-etal-2004-lpnmr,liu-trus-2005-aaai} try to maintain
subset minimality and therefore differ on this program with respect to
\cite{dell-etal-2003b,gelf-2002,eite-etal-97g}. We conclude that this program does not have a semantics which is generally agreed upon.
\end{example}

Our policy, in the development of \dlv, is to keep the system language
as much agreed-upon as possible, and to try to guarantee a clear and
intuitive semantics for the newly introduced constructs.  Thus, we
disregard programs which are not aggregate-stratified, leaving their
introduction in \dlv to future work.%
\footnote{
Note that the limitation to aggregate-stratified programs is justified also from philosophical perspectives.
For instance, defining a class $q$ before defining subsets of $q$ has
been recommended by Zermelo, but we will not go into details of this aspect.
}

In addition, we observe that unstratified aggregates may cause a computational
overhead.
For instance, the complexity of brave and cautious reasoning
on normal programs without weak constraints jumps from $\NP$ and $\CONP$ to $\SigmaP{2}$ and $\PiP{2}$,
respectively, if unstratified aggregates are allowed
\cite{ferr-2005-lpnmr,cali-etal-2005-ijcai}, while it remains in $\NP$ and $\CONP$
if aggregates are stratified.

\subsection{Comparison to the Language of Lparse}
\label{sec:lparse}

Very related to \DLPA is without doubt the language of
Lparse \cite{lparse-manual}, which serves as a grounding frontend to systems like
Smodels \cite{simo-etal-2002}, Cmodels \cite{lier-05}, or clasp \cite{gebs-etal-2007-ijcai}, which deal with
aggregates.  We observe a strong similarity between cardinality
constraints and $\countagg$, as well as weight constraints and
$\sumagg$, respectively.  Indeed, the \DLPA encodings of both Team
Building and Seating can be easily translated to the language of
Lparse.  However, there are several relevant differences.

\DLPA aggregates like $\minagg$, $\maxagg$,
and $\timesagg$ do not have a counterpart in the language of Lparse.
Moreover, \DLPA provides a general syntactic framework into which further
aggregates can be easily included.

In \DLPA aggregate atoms can be negated, while cardinality and weight
constraints in the language of Lparse cannot.  Negated aggregates are
useful for a more direct knowledge representation, and allow to
express, for instance, that some value should be external to a given
range. For example, $ \naf 3 \leq \countagg\{ X:p(X)\} \leq 7$ is true
if the number of true facts for $p$ is in $[0,3[ \cup ]7,\infty[$; for
expressing the same property in Smodels one has to write two
cardinality constraints.

Furthermore, note that symbolic sets of \DLPA directly represent pure
sets of term tuples, and by means of projection \DLPA can also
represent multisets naturally (see the discussion on {Team Building}
in Section~\ref{sec:KR}). In contrast, cardinality constraints work on
sets of ground atoms, rather than multisets of terms. For instance,
Condition $p_2$ of {Team Building} in Section~\ref{sec:KR} cannot be
directly encoded in a constraint, but needs the definition of an
auxiliary predicate. Weight constraints, on the other hand, work
exclusively on multisets of numbers, and additional rules are needed
to encode pure sets.

The language of Lparse requires that each variable has to occur
in a positive atom formed by a \emph{domain predicate} which must not
be recursive with a head atom --- by default a domain predicate must
be defined by an aggregate-free program. It follows that the language
of Lparse has no equivalent to assignment aggregates, which prohibits
the definition of simple concepts such as determining the cardinality
of input relations, as discussed in Section~\ref{sec:KR}.

The language of Lparse does however allow for cardinality and weight
constraints in the heads of rules, while \DLPA aggregates may only
occur in rule bodies. The presence of weight constraints in heads is
an interesting feature, which allows, for instance, to ``guess'' an
arbitrary subset of a given set. But it causes the loss of some
semantic properties of non-monotonic languages, see
\cite{mare-remm-2002}. Lparse rules having cardinality and weight
constraints in the head can however be expressed in \DLPA in the
following way: The atom to be aggregated over is put into a
disjunctive head, which also contains a copy of this atom in which the
predicate symbol is replaced by a fresh one, keeping the body of the
original rule augmented by the ``domain atom'' of the constraint
atom. Moreover, an integrity constraint is generated, which contains
the negated constraint atom (transformed into a corresponding \DLPA
aggregate atom) from the original rule head and the body of the
original rule. Transforming an Lparse program in this way to a \DLPA
program (replacing also cardinality and weight constraint atoms in
rule bodies by corresponding \DLPA aggregate atoms), the answer sets
of the resulting \DLPA program without atoms containing the fresh
predicates are precisely the answer sets of the original Lparse
program.

Moreover, the language of Lparse does allow for
aggregate-unstratified programs, with the intended semantics of
\cite{niem-etal-99}. As discussed earlier, there is currently no
consensus about the semantics of aggregate-unstratified programs, and
indeed the semantics of \cite{niem-etal-99} has been criticized to
yield unintuitive results when weight constraints over signed
integers are present \cite{ferr-lifs-2005}.

\subsection{Comparison to the language of \smodelsa{}}
\label{sec:smodelsa}

More recently, the system \smodelsa{} has been described in
\cite{elka-etal-2005-lpnmr}. Its language is an extension of the language of Lparse which allows for aggregates (possibly not aggregate-stratified) under the semantic described in
\cite{son-etal-2005-tr}, which coincides on the semantics of \DLPA on aggregate-stratified programs.

The syntax of the additional aggregate constructs allowed in
\smodelsa{} is more similar to the one of \DLPA (compared to those of
Lparse, which are also available in \smodelsa{}), and allows for
$sum$, $count$, $min$, $max$, and also $avg$, while $times$ is
currently not supported. In this sense, the \smodelsa{} can be
considered the system, which is most similar to \DLP.

There is, however, one rather crucial difference in the
aggregate syntax of \smodelsa{}: There may be only one term to be
aggregated over. This means that, for example, the following \DLPA rule has
no counterpart in terms of the new aggregate constructs in \smodelsa{}.
\begin{dlvcode}
tooexpensive \derives \sumagg\{Cost,Item: order(Item,Cost)\} \!> 100.
\end{dlvcode}
The intended meaning of this rule is that $tooexpensive$ should be
derived when the sum of the costs of all ordered items exceeds a
threshold of 100. Note that here we specified two terms to be
aggregated over, where the sum will be computed over the first
one. This is important, as different items may incur the same cost. For instance if
$order(valve,60)$ and $order(pipe,60)$ hold, then $tooexpensive$ should be
derived. One may try to write the following variant in the syntax of \smodelsa{}:
\begin{dlvcode}
tooexpensive \derives sum(Cost, order(Item,Cost)) \!> 100.
\end{dlvcode}
However, when $order(valve,60)$ and $order(pipe,60)$ hold,
$tooexpensive$ would not be derived, as 60 is summed only
once. Currently, there does not seem to be any way of circumventing
this problem with the aggregates introduced by \smodelsa.

Actually, there is a second problem with the rule mentioned above in
the current version of \smodelsa. The way in which the preprocessing
is done requires that each variable in the aggregate atom is domain
restricted by an atom outside the aggregate. In this rule, the
condition is not met, but it is not possible either to add an atom
outside the aggregate for domain restricting $Item$ without changing
the semantics of the rule. However, in many cases these problems can
be overcome by writing an equivalent weight constraint in the language
of Lparse, which are also available in \smodelsa.

Other differences between the language of \smodelsa{} and \DLPA are
that aggregate atoms may not occur negated, that all variables must be
domain restricted, that each rule may contain only one aggregate and
that assignment aggregates are not permitted. Moreover, there is
currently no possibility to specify a conjunction of literals (rather
than a single atom) inside an aggregate atom in \smodelsa; but one can
fairly easily circumvent this limitation by replacing the conjunction
by a new atom which is then defined by an appropriate rule.  These
differences are similar to the differences between \DLPA and the
language of Lparse. Moreover, given that \smodelsa{} relies on Smodels
as an engine, it also does not support disjunctive rules under the
semantics of \DLPA.

On the system level, the architecture of \smodelsa{} considerably differs from
the one of \dlva. It first preprocesses the input using an algorithm
implemented in Prolog, yielding an intermediate program. This program
is then submitted to Lparse. The output of Lparse is subsequently
processed by a transformation algorithm (also implemented in Prolog),
whose output is then submitted to Lparse another time. Finally, Smodels is
called on the output of the second Lparse invocation to compute the
answer sets. The key idea of the system is to compile away the
aggregates, creating new rules or constraints, which emulate the
aggregate atoms. In contrast, in the implementation of
\dlva, aggregates are first-class citizens and all the internal
algorithms of \dlv have been updated in order to deal with aggregates.

\section{Conclusion}
\label{sec:conclusion}

We have proposed \DLPA, an extension of \DLP by aggregate functions
count, sum, times, min, and max, and have implemented this in the \dlv
system.  On the one hand, we have demonstrated that the aggregate
functions increase the knowledge modeling power of \DLP, supporting a
more natural and concise knowledge representation.  On the other hand,
we have shown that aggregate functions do not increase the complexity
of the main reasoning tasks.  In fact, experiments have confirmed that
the succinctness of the encodings employing aggregates has a strong
positive impact on the efficiency of the computation.

Future work will concern the introduction of further aggregate operators
like \anyagg\ (``Is there any matching element in the set?'') and \avgagg,
investigations of a general framework that will allow adding further
aggregates much more easily, 
extending semantics to classes of programs which are not
aggregate-stratified,
as well as the design of further optimization techniques and heuristics to
improve the efficiency of the computation.

\section*{Acknowledgements}
This work has greatly benefited from interesting discussions with and
comments by Paolo Ferraris, Michael Gelfond, Vladimir Lifschitz,
Nikolay Pelov, and from the comments and suggestions by the anonymous referees.
It was partially supported by M.U.R.\ under the PRIN project
``Potenziamento e Applicazioni della Programmazione Logica Disgiuntiva'',
and by M.I.U.R.\ under internationalization project
``Sistemi basati sulla logica per la rappresentazione di conoscenza:
estensioni e tecniche di ottimizzazione''.
Wolfgang Faber's work was funded by an APART grant of the Austrian
Academy of Sciences.

\bibliography{../bibtex/bibtex}

\newcommand{\SortNoOp}[1]{}
\begin{thebibliography}{}

\bibitem[\protect\citeauthoryear{Apt, Blair, and Walker}{Apt
  et~al\mbox{.}}{1988}]{apt-etal-88}
{\sc Apt, K.~R.}, {\sc Blair, H.~A.}, {\sc and} {\sc Walker, A.} 1988.
\newblock {Towards a Theory of Declarative Knowledge}.
\newblock In {\em {Foundations of Deductive Databases and Logic Programming}},
  {J.~Minker}, Ed. Morgan Kaufmann Publishers, Inc., Washington DC, 89--148.

\bibitem[\protect\citeauthoryear{Baral}{Baral}{2003}]{bara-2002}
{\sc Baral, C.} 2003.
\newblock {\em {Knowledge Representation, Reasoning and Declarative Problem
  Solving}}.
\newblock Cambridge University Press.

\bibitem[\protect\citeauthoryear{Ben-Eliyahu and Dechter}{Ben-Eliyahu and
  Dechter}{1994}]{bene-dech-94}
{\sc Ben-Eliyahu, R.} {\sc and} {\sc Dechter, R.} 1994.
\newblock {Propositional Semantics for Disjunctive Logic Programs}.
\newblock {\em Annals of Mathematics and Artificial Intelligence\/}~{\em 12},
  53--87.

\bibitem[\protect\citeauthoryear{Buccafurri, Leone, and Rullo}{Buccafurri
  et~al\mbox{.}}{2000}]{bucc-etal-2000a}
{\sc Buccafurri, F.}, {\sc Leone, N.}, {\sc and} {\sc Rullo, P.} 2000.
\newblock {Enhancing Disjunctive Datalog by Constraints}.
\newblock {\em {IEEE Transactions on Knowledge and Data Engineering}\/}~{\em
  12,\/}~5, 845--860.

\bibitem[\protect\citeauthoryear{Calimeri, Faber, Leone, and Perri}{Calimeri
  et~al\mbox{.}}{2005}]{cali-etal-2005-ijcai}
{\sc Calimeri, F.}, {\sc Faber, W.}, {\sc Leone, N.}, {\sc and} {\sc Perri, S.}
  2005.
\newblock {Declarative and Computational Properties of Logic Programs with
  Aggregates}.
\newblock In {\em {Nineteenth International Joint Conference on Artificial
  Intelligence (IJCAI-05)}}. 406--411.

\bibitem[\protect\citeauthoryear{Chimenti, Gamboa, Krishnamurthy, Naqvi, Tsur,
  and Zaniolo}{Chimenti et~al\mbox{.}}{1990}]{chim-etal-90}
{\sc Chimenti, D.}, {\sc Gamboa, R.}, {\sc Krishnamurthy, R.}, {\sc Naqvi,
  S.~A.}, {\sc Tsur, S.}, {\sc and} {\sc Zaniolo, C.} 1990.
\newblock {The LDL System Prototype}.
\newblock {\em {IEEE Transactions on Knowledge and Data Engineering}\/}~{\em
  2,\/}~1.

\bibitem[\protect\citeauthoryear{Dell'Armi, Faber, Ielpa, Leone, and
  Pfeifer}{Dell'Armi et~al\mbox{.}}{2003}]{dell-etal-2003b}
{\sc Dell'Armi, T.}, {\sc Faber, W.}, {\sc Ielpa, G.}, {\sc Leone, N.}, {\sc
  and} {\sc Pfeifer, G.} 2003.
\newblock {Aggregate Functions in DLV}.
\newblock In {\em {Proceedings ASP03 - Answer Set Programming: Advances in
  Theory and Implementation}}, {M.~{de Vos}} {and} {A.~Provetti}, Eds. Messina,
  Italy, 274--288.
\newblock Online at \url{http://CEUR-WS.org/Vol-78/}.

\bibitem[\protect\citeauthoryear{Dowling and Gallier}{Dowling and
  Gallier}{1984}]{dowl-gall-84}
{\sc Dowling, W.~F.} {\sc and} {\sc Gallier, J.~H.} 1984.
\newblock {Linear-time Algorithms for Testing the Satisfability of
  Propositional Horn Formulae}.
\newblock {\em {Journal of Logic Programming}\/}~{\em 3}, 267--284.

\bibitem[\protect\citeauthoryear{Eiter, Faber, Leone, and Pfeifer}{Eiter
  et~al\mbox{.}}{2000}]{eite-etal-2000c}
{\sc Eiter, T.}, {\sc Faber, W.}, {\sc Leone, N.}, {\sc and} {\sc Pfeifer, G.}
  2000.
\newblock {Declarative Problem-Solving Using the DLV System}.
\newblock In {\em {Logic-Based Artificial Intelligence}}, {J.~Minker}, Ed.
  Kluwer Academic Publishers, 79--103.

\bibitem[\protect\citeauthoryear{Eiter and Gottlob}{Eiter and
  Gottlob}{1995}]{eite-gott-95}
{\sc Eiter, T.} {\sc and} {\sc Gottlob, G.} 1995.
\newblock {On the Computational Cost of Disjunctive Logic Programming:
  Propositional Case}.
\newblock {\em {Annals of Mathematics and Artificial Intelligence}\/}~{\em
  15,\/}~3/4, 289--323.

\bibitem[\protect\citeauthoryear{Eiter, Gottlob, and Mannila}{Eiter, Gottlob, and Mannila}{1997}]{eite-etal-97f}
{\sc Eiter, T.}, {\sc Gottlob, G.}, {\sc and} {\sc Mannila, H.} 1997.
\newblock {Disjunctive Datalog}.
\newblock {\em {ACM Transactions on Database Systems}\/}~{\em 22,\/}~3 (Sept.),
  364--418.

\bibitem[\protect\citeauthoryear{Eiter, Gottlob, and Veith}{Eiter, Gottlob, and Veith}{1997}]{eite-etal-97g}
{\sc Eiter, T.}, {\sc Gottlob, G.}, {\sc and} {\sc Veith, H.} 1997.
\newblock {Modular Logic Programming and Generalized Quantifiers}.
\newblock In {\em {Proceedings of the 4th International Conference on Logic
  Programming and Nonmonotonic Reasoning (LPNMR-97)}}, {J.~Dix}, {U.~Furbach},
  {and} {A.~Nerode}, Eds. LNCS, vol. 1265. Springer, 290--309.

\bibitem[\protect\citeauthoryear{Elkabani, Pontelli, and Son}{Elkabani
  et~al\mbox{.}}{2005}]{elka-etal-2005-lpnmr}
{\sc Elkabani, I.}, {\sc Pontelli, E.}, {\sc and} {\sc Son, T.~C.} 2005.
\newblock {\sc Smodels$^{A}$}---a system for computing answer sets of logic
  programs.
\newblock In {\em {Logic Programming and Nonmonotonic Reasoning --- 8th
  International Conference, LPNMR'05, Diamante, Italy, September 2005,
  Proceedings}}, {C.~Baral}, {G.~Greco}, {N.~Leone}, {and} {G.~Terracina}, Eds.
  {Lecture Notes in Computer Science}, vol. 3662. Springer Verlag, 427--431.

\bibitem[\protect\citeauthoryear{Faber}{Faber}{2002}]{fabe-2002}
{\sc Faber, W.} 2002.
\newblock {Enhancing Efficiency and Expressiveness in Answer Set Programming
  Systems}.
\newblock Ph.D. thesis, {Institut f{\"u}r Informationssysteme, Technische
  Universit{\"a}t Wien}.

\bibitem[\protect\citeauthoryear{Faber}{Faber}{2005}]{fabe-2005-lpnmr}
{\sc Faber, W.} 2005.
\newblock {Unfounded Sets for Disjunctive Logic Programs with Arbitrary
  Aggregates}.
\newblock In {\em {Logic Programming and Nonmonotonic Reasoning --- 8th
  International Conference, LPNMR'05, Diamante, Italy, September 2005,
  Proceedings}}, {C.~Baral}, {G.~Greco}, {N.~Leone}, {and} {G.~Terracina}, Eds.
  {Lecture Notes in Computer Science}, vol. 3662. Springer Verlag, 40--52.

\bibitem[\protect\citeauthoryear{Faber, Leone, Mateis, and Pfeifer}{Faber, Leone, Mateis, and Pfeifer}{1999}]{fabe-etal-99c}
{\sc Faber, W.}, {\sc Leone, N.}, {\sc Mateis, C.}, {\sc and} {\sc Pfeifer, G.}
  1999.
\newblock {Using Database Optimization Techniques for Nonmonotonic Reasoning}.
\newblock In {\em Proceedings of the 7th International Workshop on Deductive
  Databases and Logic Programming (DDLP'99)}, {{INAP Organizing Committee}},
  Ed. Prolog Association of Japan, 135--139.

\bibitem[\protect\citeauthoryear{Faber, Leone, and Pfeifer}{Faber, Leone, and Pfeifer}{1999}]{fabe-etal-99b}
{\sc Faber, W.}, {\sc Leone, N.}, {\sc and} {\sc Pfeifer, G.} 1999.
\newblock {Pushing Goal Derivation in DLP Computations}.
\newblock In {\em {Proceedings of the 5th International Conference on Logic
  Programming and Nonmonotonic Reasoning (LPNMR'99)}}, {M.~Gelfond},
  {N.~Leone}, {and} {G.~Pfeifer}, Eds. {Lecture Notes in AI (LNAI)}, vol. 1730.
  Springer Verlag, El Paso, Texas, USA, 177--191.

\bibitem[\protect\citeauthoryear{Faber, Leone, and Pfeifer}{Faber
  et~al\mbox{.}}{2001}]{fabe-etal-2001a}
{\sc Faber, W.}, {\sc Leone, N.}, {\sc and} {\sc Pfeifer, G.} 2001.
\newblock {Experimenting with Heuristics for Answer Set Programming}.
\newblock In {\em {Proceedings of the Seventeenth International Joint
  Conference on Artificial Intelligence (IJCAI) 2001}}. Morgan Kaufmann
  Publishers, Seattle, WA, USA, 635--640.

\bibitem[\protect\citeauthoryear{Faber, Leone, and Pfeifer}{Faber
  et~al\mbox{.}}{2004}]{fabe-etal-2004-jelia}
{\sc Faber, W.}, {\sc Leone, N.}, {\sc and} {\sc Pfeifer, G.} 2004.
\newblock Recursive aggregates in disjunctive logic programs: Semantics and
  complexity.
\newblock In {\em {Proceedings of the 9th European Conference on Artificial
  Intelligence (JELIA 2004)}}, {J.~J. Alferes} {and} {J.~Leite}, Eds. {Lecture
  Notes in AI (LNAI)}, vol. 3229. Springer Verlag, 200--212.

\bibitem[\protect\citeauthoryear{Ferraris}{Ferraris}{2005}]{ferr-2005-lpnmr}
{\sc Ferraris, P.} 2005.
\newblock {Answer Sets for Propositional Theories}.
\newblock In {\em {Logic Programming and Nonmonotonic Reasoning --- 8th
  International Conference, LPNMR'05, Diamante, Italy, September 2005,
  Proceedings}}, {C.~Baral}, {G.~Greco}, {N.~Leone}, {and} {G.~Terracina}, Eds.
  {Lecture Notes in Computer Science}, vol. 3662. Springer Verlag, 119--131.

\bibitem[\protect\citeauthoryear{Ferraris and Lifschitz}{Ferraris and
  Lifschitz}{2005}]{ferr-lifs-2005}
{\sc Ferraris, P.} {\sc and} {\sc Lifschitz, V.} 2005.
\newblock {Weight constraints as nested expressions}.
\newblock {\em {Theory and Practice of Logic Programming}\/}~{\em 5,\/}~1--2,
  45--74.

\bibitem[\protect\citeauthoryear{Gebser, Kaufmann, Neumann, and Schaub}{Gebser
  et~al\mbox{.}}{2007}]{gebs-etal-2007-ijcai}
{\sc Gebser, M.}, {\sc Kaufmann, B.}, {\sc Neumann, A.}, {\sc and} {\sc Schaub,
  T.} 2007.
\newblock Conflict-driven answer set solving.
\newblock In {\em {Twentieth International Joint Conference on Artificial
  Intelligence (IJCAI-07)}}. Morgan Kaufmann Publishers, 386--392.

\bibitem[\protect\citeauthoryear{Gebser, Liu, Namasivayam, Neumann, Schaub, and
  Truszczy{\'n}ski}{Gebser
  et~al\mbox{.}}{2007}]{gebs-etal-2007-lpnmr-competition}
{\sc Gebser, M.}, {\sc Liu, L.}, {\sc Namasivayam, G.}, {\sc Neumann, A.}, {\sc
  Schaub, T.}, {\sc and} {\sc Truszczy{\'n}ski, M.} 2007.
\newblock The first answer set programming system competition.
\newblock In {\em {Logic Programming and Nonmonotonic Reasoning --- 9th
  International Conference, LPNMR'07}}, {C.~Baral}, {G.~Brewka}, {and}
  {J.~Schlipf}, Eds. {Lecture Notes in Computer Science}, vol. 4483. Springer
  Verlag, Tempe, Arizona, 3--17.

\bibitem[\protect\citeauthoryear{Gelfond}{Gelfond}{2002}]{gelf-2002}
{\sc Gelfond, M.} 2002.
\newblock {Representing Knowledge in A-Prolog}.
\newblock In {\em {Computational Logic. Logic Programming and Beyond}}, {A.~C.
  Kakas} {and} {F.~Sadri}, Eds. LNCS, vol. 2408. Springer, 413--451.

\bibitem[\protect\citeauthoryear{Gelfond and Lifschitz}{Gelfond and
  Lifschitz}{1991}]{gelf-lifs-91}
{\sc Gelfond, M.} {\sc and} {\sc Lifschitz, V.} 1991.
\newblock {Classical Negation in Logic Programs and Disjunctive Databases}.
\newblock {\em {New Generation Computing}\/}~{\em 9}, 365--385.

\bibitem[\protect\citeauthoryear{Gottlob, Leone, and Veith}{Gottlob
  et~al\mbox{.}}{1999}]{gott-etal-99apal}
{\sc Gottlob, G.}, {\sc Leone, N.}, {\sc and} {\sc Veith, H.} 1999.
\newblock {Succinctness as a Source of Expression Complexity}.
\newblock {\em {Annals of Pure and Applied Logic}\/}~{\em 97,\/}~1--3,
  231--260.

\bibitem[\protect\citeauthoryear{Hella, Libkin, Nurmonen, and Wong}{Hella
  et~al\mbox{.}}{2001}]{hell-etal-2001-jacm}
{\sc Hella, L.}, {\sc Libkin, L.}, {\sc Nurmonen, J.}, {\sc and} {\sc Wong, L.}
  2001.
\newblock Logics with aggregate operators.
\newblock {\em {Journal of the ACM}\/}~{\em 48,\/}~4, 880--907.

\bibitem[\protect\citeauthoryear{Kemp and Ramamohanarao}{Kemp and
  Ramamohanarao}{1998}]{kemp-rama-1998}
{\sc Kemp, D.~B.} {\sc and} {\sc Ramamohanarao, K.} 1998.
\newblock {Efficient Recursive Aggregation and Negation in Deductive
  Databases}.
\newblock {\em {IEEE Transactions on Knowledge and Data Engineering}\/}~{\em
  10}, 727--745.

\bibitem[\protect\citeauthoryear{Leone, Perri, and Scarcello}{Leone
  et~al\mbox{.}}{2001}]{leon-etal-2001a}
{\sc Leone, N.}, {\sc Perri, S.}, {\sc and} {\sc Scarcello, F.} 2001.
\newblock {Improving ASP Instantiators by Join-Ordering Methods}.
\newblock In {\em {Logic Programming and Nonmonotonic Reasoning --- 6th
  International Conference, LPNMR'01, Vienna, Austria}}, {T.~Eiter},
  {W.~Faber}, {and} {M.~Truszczy\'nski}, Eds. {Lecture Notes in AI (LNAI)},
  vol. 2173. Springer Verlag, 280--294.

\bibitem[\protect\citeauthoryear{Leone, Pfeifer, Faber, Eiter, Gottlob, Perri,
  and Scarcello}{Leone et~al\mbox{.}}{2006}]{leon-etal-2002-dlv}
{\sc Leone, N.}, {\sc Pfeifer, G.}, {\sc Faber, W.}, {\sc Eiter, T.}, {\sc
  Gottlob, G.}, {\sc Perri, S.}, {\sc and} {\sc Scarcello, F.} 2006.
\newblock {The DLV System for Knowledge Representation and Reasoning}.
\newblock {\em {ACM Transactions on Computational Logic}\/}~{\em 7,\/}~3
  (July), 499--562.

\bibitem[\protect\citeauthoryear{Leone, Rullo, and Scarcello}{Leone
  et~al\mbox{.}}{1997}]{leon-etal-97b}
{\sc Leone, N.}, {\sc Rullo, P.}, {\sc and} {\sc Scarcello, F.} 1997.
\newblock {Disjunctive Stable Models: Unfounded Sets, Fixpoint Semantics and
  Computation}.
\newblock {\em {Information and Computation}\/}~{\em 135,\/}~2 (June), 69--112.

\bibitem[\protect\citeauthoryear{Lierler}{Lierler}{2005}]{lier-05}
{\sc Lierler, Y.} 2005.
\newblock {Cmodels for Tight Disjunctive Logic Programs.}
\newblock In {\em W(C)LP 19th Workshop on (Constraint) Logic Programming, Ulm,
  Germany}. Ulmer Informatik-Berichte. Universit{\"a}t Ulm, Germany, 163--166.

\bibitem[\protect\citeauthoryear{Liu and Truszczy{\'n}ski}{Liu and
  Truszczy{\'n}ski}{2005}]{liu-trus-2005-aaai}
{\sc Liu, L.} {\sc and} {\sc Truszczy{\'n}ski, M.} 2005.
\newblock Properties of programs with monotone and convex constraints.
\newblock In {\em Proceedings of the 20th National Conference on Artificial
  Intelligence (AAAI'05)}, {M.~M. Veloso} {and} {S.~Kambhampati}, Eds.
  701--706.

\bibitem[\protect\citeauthoryear{Marek, Niemel{\"a}, and
  Truszczy{\'n}ski}{Marek et~al\mbox{.}}{2004}]{mare-etal-2004-lpnmr}
{\sc Marek, V.~W.}, {\sc Niemel{\"a}, I.}, {\sc and} {\sc Truszczy{\'n}ski, M.}
  2004.
\newblock {Logic Programming with Monotone Cardinality Atom}.
\newblock In {\em {Proceedings of the 7th International Conference on Logic
  Programming and Non-Monotonic Reasoning (LPNMR-7)}}, {V.~Lifschitz} {and}
  {I.~Niemel{\"a}}, Eds. LNAI, vol. 2923. Springer, 154--166.

\bibitem[\protect\citeauthoryear{Marek and Remmel}{Marek and
  Remmel}{2002}]{mare-remm-2002}
{\sc Marek, V.~W.} {\sc and} {\sc Remmel, J.~B.} 2002.
\newblock {On Logic Programs with Cardinality Constraints}.
\newblock In {\em {Proceedings of the 9th International Workshop on
  Non-Monotonic Reasoning (NMR'2002)}}, {S.~Benferhat} {and} {E.~Giunchiglia},
  Eds. Toulouse, France, 219--228.

\bibitem[\protect\citeauthoryear{Marek and Truszczy{\'n}ski}{Marek and
  Truszczy{\'n}ski}{1991}]{mare-trus-91}
{\sc Marek, V.~W.} {\sc and} {\sc Truszczy{\'n}ski, M.} 1991.
\newblock {Autoepistemic Logic}.
\newblock {\em {Journal of the ACM}\/}~{\em 38,\/}~3, 588--619.

\bibitem[\protect\citeauthoryear{Minker}{Minker}{1982}]{mink-82}
{\sc Minker, J.} 1982.
\newblock {On Indefinite Data Bases and the Closed World Assumption}.
\newblock In {\em {Proceedings $6^{th}$ Conference on Automated Deduction (CADE
  '82)}}, {D.~W. Loveland}, Ed. {Lecture Notes in Computer Science}, vol. 138.
  Springer, New York, 292--308.

\bibitem[\protect\citeauthoryear{Minoux}{Minoux}{1988}]{mino-88}
{\sc Minoux, M.} 1988.
\newblock {LTUR: A Simplified Linear-time Unit Resolution Algorithm for Horn
  Formulae and Computer Implementation}.
\newblock {\em {Information Processing Letters}\/}~{\em 29}, 1--12.

\bibitem[\protect\citeauthoryear{Niemel{\"a}, Simons, and Soininen}{Niemel{\"a}
  et~al\mbox{.}}{1999}]{niem-etal-99}
{\sc Niemel{\"a}, I.}, {\sc Simons, P.}, {\sc and} {\sc Soininen, T.} 1999.
\newblock {Stable Model Semantics of Weight Constraint Rules}.
\newblock In {\em {Proceedings of the 5th International Conference on Logic
  Programming and Nonmonotonic Reasoning (LPNMR'99)}}, {M.~Gelfond},
  {N.~Leone}, {and} {G.~Pfeifer}, Eds. {Lecture Notes in AI (LNAI)}, vol. 1730.
  Springer Verlag, El Paso, Texas, USA, 107--116.

\bibitem[\protect\citeauthoryear{Pelov}{Pelov}{2004}]{pelo-2004}
{\sc Pelov, N.} 2004.
\newblock {Semantics of Logic Programs with Aggregates}.
\newblock Ph.D. thesis, Katholieke Universiteit Leuven, Leuven, Belgium.

\bibitem[\protect\citeauthoryear{Pelov, Denecker, and Bruynooghe}{Pelov
  et~al\mbox{.}}{2004}]{pelo-etal-2004a}
{\sc Pelov, N.}, {\sc Denecker, M.}, {\sc and} {\sc Bruynooghe, M.} 2004.
\newblock Partial stable models for logic programs with aggregates.
\newblock In {\em {Proceedings of the 7th International Conference on Logic
  Programming and Non-Monotonic Reasoning (LPNMR-7)}}. {Lecture Notes in AI
  (LNAI)}, vol. 2923. Springer, 207--219.

\bibitem[\protect\citeauthoryear{Przymusinski}{Przymusinski}{1988}]{przy-88}
{\sc Przymusinski, T.~C.} 1988.
\newblock {On the Declarative Semantics of Deductive Databases and Logic
  Programs}.
\newblock In {\em {Foundations of Deductive Databases and Logic Programming}},
  {J.~Minker}, Ed. Morgan Kaufmann Publishers, Inc., 193--216.

\bibitem[\protect\citeauthoryear{Ross and Sagiv}{Ross and
  Sagiv}{1997}]{ross-sagi-1997}
{\sc Ross, K.~A.} {\sc and} {\sc Sagiv, Y.} 1997.
\newblock {Monotonic Aggregation in Deductive Databases}.
\newblock {\em {Journal of Computer and System Sciences}\/}~{\em 54,\/}~1
  (Feb.), 79--97.

\bibitem[\protect\citeauthoryear{Simons, Niemel{\"a}, and Soininen}{Simons
  et~al\mbox{.}}{2002}]{simo-etal-2002}
{\sc Simons, P.}, {\sc Niemel{\"a}, I.}, {\sc and} {\sc Soininen, T.} 2002.
\newblock {Extending and Implementing the Stable Model Semantics}.
\newblock {\em {Artificial Intelligence}\/}~{\em 138}, 181--234.

\bibitem[\protect\citeauthoryear{Son and Pontelli}{Son and
  Pontelli}{2007}]{son-pont-2007-tplp}
{\sc Son, T.~C.} {\sc and} {\sc Pontelli, E.} 2007.
\newblock {A Constructive Semantic Characterization of Aggregates in ASP}.
\newblock {\em {Theory and Practice of Logic Programming}\/}~{\em 7}, 355--375.

\bibitem[\protect\citeauthoryear{Son, Pontelli, and Elkabani}{Son
  et~al\mbox{.}}{2005}]{son-etal-2005-tr}
{\sc Son, T.~C.}, {\sc Pontelli, E.}, {\sc and} {\sc Elkabani, I.} 2005.
\newblock {On Logic Programming with Aggregates}.
\newblock Tech. Rep. NMSU-CS-2005-006, New Mexico State University.

\bibitem[\protect\citeauthoryear{Syrj{\"a}nen}{Syrj{\"a}nen}{2002}]{lparse-man%
ual}
{\sc Syrj{\"a}nen, T.} 2002.
\newblock {Lparse 1.0 User's Manual}.
\newblock \url{http://www.tcs.hut.fi/Software/smodels/lparse.ps.gz}.

\bibitem[\protect\citeauthoryear{Ullman}{Ullman}{1989}]{ullm-89}
{\sc Ullman, J.~D.} 1989.
\newblock {\em {Principles of Database and Knowledge Base Systems}}.
\newblock Computer Science Press.

\end{thebibliography}
\bibliographystyle{acmtrans}

\end{document}